\documentclass[journal]{IEEEtran}
\PassOptionsToPackage{table}{xcolor}

\usepackage{cite}
\usepackage{textcomp}

\usepackage{amsmath,amssymb}
\usepackage{algorithmic}
\usepackage{graphicx}
\usepackage{color}
\usepackage{booktabs}
\usepackage{makecell}
\usepackage{multirow}
\usepackage{mathrsfs}
\usepackage{rotating}
\usepackage{balance}
\usepackage{cite}
\usepackage{ulem}
\usepackage{multicol}
\usepackage{amssymb}
\usepackage{xcolor}
\definecolor{wrong}{rgb}{.8,.349,.1}
\definecolor{right}{rgb}{.3,.7,.1}
\usepackage{newfloat}
\usepackage{listings}
\usepackage{colortbl}
\usepackage{bm}

\usepackage{threeparttable}
\usepackage{dsfont}

\usepackage{subfigure}

\DeclareGraphicsExtensions{.png,.png,.jpg,.fig,.tif}
\definecolor{color1}{rgb}{0.2353,0.2353,1}
\definecolor{color2}{rgb}{0.8000,0,0.4000}
\definecolor{color3}{rgb}{0.6980,0.4000,1}
\definecolor{color4}{rgb}{1,0.6,1}

\newcolumntype{L}[1]{>{\raggedright\arraybackslash}p{#1}}
\newcolumntype{C}[1]{>{\centering\arraybackslash}p{#1}}
\newcolumntype{R}[1]{>{\raggedleft\arraybackslash}p{#1}}

\def\BibTeX{{\rm B\kern-.05em{\sc i\kern-.025em b}\kern-.08em
    T\kern-.1667em\lower.7ex\hbox{E}\kern-.125emX}}
\begin{document}
\title{Diverse Teaching and Label Propagation  for Generic Semi-Supervised Medical Image Segmentation}
% $^{\ast,\dagger}$
\author{Wei Li, Pengcheng Zhou, Linye Ma, Wenyi Zhao, Huihua Yang, Yuchen Guo$^\dagger$ 
% \author{Wei Li, Pengcheng Zhou, Linye Ma, Wenyi Zhao$^{\dagger}$, Huihua Yang
% \thanks{$^\ast$ indicates equal contribution, $\dagger$ indicates corresponding author.}
\thanks{$\dagger$ indicates corresponding author.}
\thanks{This research was supported in part by the  Beijing-Tianjin-Hebei Natural Science Foundation Cooperation Project (No. H2024102009), and in part by the National Natural Science Foundation of China (No. 62376038), and in part by the National Science and Technology Major Project (No.2023ZD0506304), and in part by the Natural Science Foundation of China (No.82441013). Corresponding author: Yuchen Guo.} 
\thanks{Wei Li and Yuchen Guo are with the Beijing Information Science and Technology National Research Center (BNRIST), Tsinghua University, Beijing, 100876, China. Pengcheng Zhou is the  with National University of Singapore College of Design and Engineering‌. Linye Ma is with The 15th Institute of China Electronics Technology Group Corporation, Beijing, 100876, China. Wenyi Zhao and Huihua Yang are with the School of Intelligent Engineering and Automation, Beijing University of Posts and Telecommunications, Beijing, 100876, China. (e-mail: leesoon@mail.tsinghua.edu.cn,yuchen.w.guo@gmail.com).}
}

\maketitle
\begin{abstract}

Limited annotation and domain shift severely limit medical image segmentation in clinical practice, necessitating unified approaches for semi-supervised segmentation (SSMIS), unsupervised domain adaptation (UMDA), and semi-supervised domain generalization (Semi-MDG). Existing methods are generally tailored to specific tasks and fail to generalize across these tasks due to error accumulation from noisy pseudo-labels and poor exploitation of data structure. In this paper, we employ a Diverse Teaching and Label Propagation  Network (DTLP-Net) to boost the generic semi-supervised medical image segmentation.  DTLP-Net involves a single student model and two diverse teacher models, where the first teacher decouples the labeled and unlabeled with a diffusion decoder, and the second is the mean-teacher model. Their outputs are then synergistically fused via an entropy-based ensemble to yield robust supervisory signals. The framework's performance is further bolstered by three synergistic consistency  strategy. First, a global-local consistency module leverages cross-set CutMix and masked image modeling to learn domain-invariant representations from both inter- and intra-sample contexts. Second, masked reconstruction on the feature level and knowledge distillation from the soft prediction is further utilized to alleviate the adverse impact of the noise present the hard pseudo labels. Finally, a voxel-level label propagation strategy explicitly models and enforces pairwise dependencies, enhancing spatial coherence in the final segmentation. We evaluate our proposed framework on five benchmark datasets for SSMIS, UMDA, and Semi-MDG tasks. The results showcase notable improvements compared to state-of-the-art methods across all five settings, indicating the potential of our framework to tackle more challenging SSL scenarios.

\end{abstract}

\section{Introduction}
\label{sec.introduction}

\IEEEPARstart{S}emi-supervised medical image segmentation (SSMIS) methods~\cite{bai2023bidirectional,li2024diversity,li2022collaborative,zhang2024interteach,chi2025cross} has emerged as a cornerstone in medical image segmentation, offering a potent solution to the prohibitive cost and expertise required for the manual labeling process. 
However, the practical utility of conventional SSMIS is frequently undermined by a second, equally critical challenge: domain shift. This phenomenon, arising from variations in scanners, patient populations, and imaging protocols, causes severe performance degradation when models are deployed in real-world clinical settings. The confluence of these two challenges has given rise to more complex and realistic scenarios, including unsupervised domain adaptation (UMDA)~\cite{zhu2017cyclegan_uda,zheng2024dual,li2025leveraging}, where the model must adapt from a labeled source domain to an entirely unlabeled target domain, and the even more demanding semi-supervised domain generalization (Semi-MDG)\cite{yao2022epl,liu2022vmfnet}. The latter aims to train a model on a collection of partially labeled source domains to perform robustly on a completely unseen domain, representing a critical step toward true clinical applicability and robustness.

Despite their shared conceptual roots in learning from limited labels, these scenarios have historically been tackled in isolation. This fragmented approach has led to a proliferation of task-specific models that lack versatility; a state-of-the-art SSMIS method may fail entirely when faced with the domain shifts inherent to UMDA or Semi-MDG~\cite{wang2021annotation,wang2024towards}.  While recent efforts like AIDE~\cite{wang2021annotation} and A\&D~\cite{wang2024towards} have pioneered a unified approach, their performance gains in cross-domain settings remain incremental. A\&D~\cite{wang2024towards} introduced an aggregating and decoupling approach to solve these three challenging tasks simultaneously; its advancements in tasks involving domain shift, \textit{e.g.}, UMDA and Semi-MDG, remain limited compared to prior art methods that are specifically designed for these tasks. We argue this is due to two fundamental, unresolved issues: (1) The generation of unreliable pseudo-labels, which become particularly noisy and damaging in the presence of domain shift, leading to error accumulation~\cite{bias2019} and model collapse~\cite{ma2023enhanced}. The main challenge in SSMIS lies in efficiently leveraging unlabeled images, while UMDA and Semi-MDG scenarios further require addressing the domain shift issues. All of these three tasks benefit from the reliable pseudo-label and model diversity~\cite{li2024diversity}.  (2) The under-utilization of intrinsic data structure, where voxel-level relationships and spatial context are not fully exploited to guide robust, domain-invariant feature learning. A\&D~\cite{wang2024towards} adopts the Diffusion-based V-Net to capture the domain-invariant features, which is not enough to learn the common knowledge of different domains.

To address this challenge, we propose a \textbf{D}iverse \textbf{T}eaching and \textbf{L}abel \textbf{P}ropagation Network (DTLP-Net) to boost the Generic Semi-Supervised Medical Image Segmentation, as shown in  see Fig.~\ref{fig:main}. The core of our DTLP-Net is a sophisticated dual-teacher strategy engineered to generate reliable pseudo-labels, even in the presence of significant domain shift. Specifically, our approach builds on A\&D~\cite{wang2024towards}, which features a shared diffusion encoder for domain-invariant feature extraction and a multi-decoder architecture (one Diffusion VNet, two V-Nets) to decouple the training process. For the unlabeled data stream, our diverse teaching mechanism generates high-quality supervision through two distinct teachers. The first teacher's predictions are derived from the decoupled decoders, while the second is generated by a EMA mean-teacher~\cite{li2024diversity}. We then employ an entropy-based ensembling strategy to intelligently fuse their outputs, encouraging the student to learn from both consistent and conflicting predictions, as illustrated in Fig.~\ref{fig:main} (f). This cross-supervision compels the decoders to learn complementary features and correct each other's mistakes, enhancing overall performance.  Moreover, DTLP-Net incorporates three powerful consistency learning mechanisms. First, we leverage a cross-set Cut-Mix strategy and masked image modeling to enforce both global and local consistency, compelling the model to learn domain-agnostic structural features. Then, since the hard pseudo labels generated by the dual teachers still inevitably contain noise, especially in the scenario where there is class imbalance, we introduce additional supervision for the student. This includes masked reconstruction, which aligns the student's predictions with the second mean-teacher's logits, and knowledge distillation from the first teacher's soft labels.  Finally, we introduce a voxel-level label propagation module that explicitly models and enforces pairwise spatial similarities within an image, enhancing segmentation coherence and robustness. By synergizing diverse teaching with multi-level consistency, our framework effectively mitigates domain shifts and maximizes information extraction from unlabeled data.  The overall contributions can be summarized as follows:
\vspace{-4mm}
\begin{itemize}
    \item We propose an effective and unified framework DTLP-Net to boost the Generic Semi-Supervised Medical Image Segmentation, including SSMIS, UMDA, and Semi-MDG tasks.
    \item We introduce a novel dual-teacher, entropy-based ensemble strategy that generates high-quality pseudo-labels in the presence of domain shift. By explicitly leveraging teacher diversity and conflict, our method effectively reduces confirmation bias. 
     \item We design a hierarchical consistency mechanism that learns domain-invariant features across multiple scales. It enforces global-local structural priors through Cut-Mix and masking, while simultaneously ensuring fine-grained spatial coherence via a novel label propagation module. Moreover, we propose masked reconstruction and knowledge distillation to mitigate the adverse impact of the noise present in the hard pseudo labels.
    \item We evaluate our approach on five standard benchmark datasets, the experimental results show that our approach attains substantial enhancements across all three scenarios of SSMIS, UMDA, and Semi-MDG.
\end{itemize}

\section{Related Works}
\label{sec.related_works}
  
\subsection{Semi-supervised medical image segmentation}

In an effort to reduce the costs related to volume-level annotation while still ensuring high accuracy, diverse semi-supervised methods for medical image segmentation tasks have been developed. Drawing inspiration from semi-supervised image classification~\cite{sohn2020fixmatch}, techniques such as self-training and consistency regularization have emerged.  In addition, the mean-teacher approach~\cite{tarvainen2017mean} and its extensions~\cite{yu2019uamt, bai2023bidirectional, guo2025optimal} have gained significant attention. For instance,  BCP~\cite{bai2023bidirectional}  advocates a straightforward Mean Teacher architecture to copy and paste annotated and unlabeled data bidirectionally, highlighting a key challenge in semi-supervised learning where the distributions of labeled and unlabeled data may diverge. Moreover, some works resort to the co-training paradigm, where multiple models are collaboratively trained by cooperatively training multiple models. CTCL~\cite{li2022collaborative} proposes a collaborative transformer-CNN learning for semi-supervised medical image segmentation.  W2sPC~\cite{yang2025semi} further incorporates weak-to-strong perturbation consistency and edge-aware contrastive representation, thus promoting the consistency of reliable regions among diverse predictions and facilitating the learning of class-discriminative representations. GapMatch~\cite{huang2025gapmatch} effectively links instance and model perturbations, thereby expanding the perturbation space. Moreover, it utilizes dual perturbation to enforce consistency regularization on the model. Different from these works that only consider one or two perturbations to enhance the model diversity, CMMT-Net~\cite{li2024diversity} proposes a unified framework to generate reliable pseudo-labels and achieve consistency learning. In contrast to the aforementioned techniques, we adopt co-training by multi-decoder to enlarge the  diversity of the models while designing multi-teachers to enhance the  pseudo-labels quality. This simple yet effective approach improves the model's generalization ability and segmentation accuracy. 

% OTCMC \cite{guo2025optimal} puts forward the utilization of Optimal Transport for the purpose of attaining efficient label propagation. Additionally, it employs Central Moment Consistency Regularization to direct the network's focus towards the geometric structure of images. 

% MC-Net+~\cite{wu2022mutual} addresses inconsistency  through mutual consistency.

% \vspace{-4mm}
\subsection{Unsupervised Medical Domain Adaptation}

Unsupervised Domain Adaptation (UDA)~\cite{zhu2017cyclegan_uda,hoffman2018cycada_uda}  aims to solve the domain shifts by jointly training the model with labeled source domain data and unlabeled target domain data. In the field of Unsupervised medical Domain Adaptation (UMDA),  one modality usually lacks any segmentation annotation, which has recently gained increasing attention since it offers an efficient way to compensate for limited medical image data.  Existing  UMDA approaches resort to various directions, \textit{e.g.} semi-supervised learning~\cite{liu2022act_uda_new}, generative adversarial-based methods~\cite{dou2019pnp_uda,han2021dsan_uda}, and contrastive learning~\cite{gu2022confuda_uda_new}, aiming to mitigate the adversarial influence caused by the severe domain shifts  via image-level~\cite{zhu2017cyclegan_uda}, feature-level~\cite{han2021dsan_uda} or both~\cite{hoffman2018cycada_uda,zheng2024dual}.  Although with promising adaptation results, these methods fail to comprehensively exploit the unlabeled target domain information, which impedes generalization. Moreover, the output-level alignment across different domains still awaits exploration.

% overlook the labeled source domain information and do not fully explore the  unlabeled target domain information, hindering the generalization. In addition, the output-level alignment of different domains remains to explore. 

\begin{figure*}[!htbp]
	\begin{center}
		%\centering
		\includegraphics[width=1.0\textwidth]{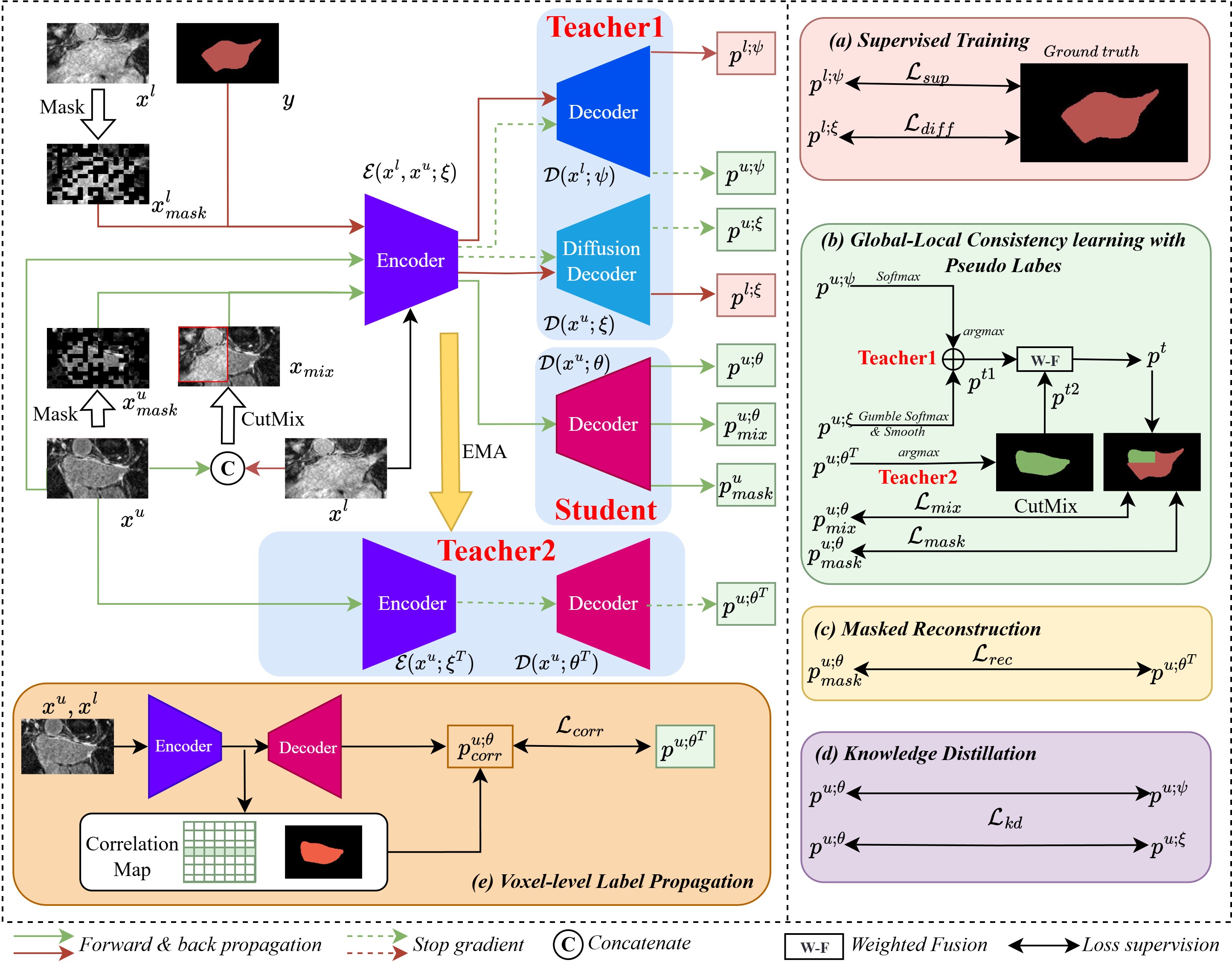}
		%\centering
		\caption{
			An illustration of our Diverse Teaching and Label Propagation Network (DTLP-Net). The training process of the decoders using labeled data and unlabeled data is decoupled. During inference, only the decoder $\mathcal{E}(x^u;\theta)$ is employed. 
}
\vspace{-5mm}
    \label{fig:main} 
	\end{center}
\end{figure*}

\vspace{-4mm}
\subsection{Semi-supervised Medical Domain Generalization}

Semi-supervised domain generalization (SemiDG) is a more challenging setting, where the model does not use any information from the target domain, and the data in the source domains are only partially labeled. Compared with SSMIS and UMDA tasks, SemiDG necessitates the model to possess robust feature extraction capabilities and a high degree of generalizability. Existing SemiDG methods~\cite{liu2021dgnet,liu2021SDNet} use various carefully designed strategies to solve the domain shifts, \textit{e.g.}, meta-learning~\cite{liu2021dgnet}, and Fourier transformation~\cite{yao2022epl}, which are not general and have unsatisfactory performance on the tasks such as UDA and SSL. Considering all of these three tasks involve learning from both labeled and unlabeled data, A\&D~\cite{wang2024towards} first try to unified with a generic framework handling all of them. However, despite the advancement they achieved, the improvement on SSMIS, UMDA and Semi-MDG tasks is marginal compared with task-tailored SOTAs.

\section{Methodology}
\label{sec.methodology}
In the 3D Generic Semi-Supervised segmentation setting, the training data comprises $N_L$ labeled samples $D_l=\left\{\left(x_i, y_i \right)\right\}_{i=1}^{N_L}$ and  $N_U$ unlabeled samples $D_u=\left\{x_i \right\}_{i=1}^{N_U}$, respectively, where $x_i \subset \mathbb{R}^{L \times W\times H}$ depicts the input volume with dimensions $L\times W \times H$, and $y_i \subset \left\{ 0, 1\right\}^{Y\times L\times H \times W}$ represents the ground truth with $Y$ classes. The aim is to utilize the datasets $D_l$ and $D_u$ to train a model to generate meaningful semantic predictions. Our DTLP-Net consists of a shared encoder $\mathcal{E}(x^l,x^u;\xi)$ to learn domain-invariant  features, and three decoders,  \textit{e.g.}, V-Net Decoder $\mathcal{D}(x^l;\psi)$, Diffusion Decoder  $\mathcal{D}(x^l,x^u;\xi)$ and V-Net Decoder $\mathcal{D}(x^u;\theta)$, where $\psi$, $\xi$, and $\theta$ are model parameters, as illustrated in Fig.~\ref{fig:main}. In addition, for  $\mathcal{D}(x^u;\theta)$ Decoder branch, we further obtain the mean-teacher Encoder $\mathcal{E}(x^l,x^u;\xi^{T})$ and  Decoder $\mathcal{D}(x^u;\theta^{T})$.

\vspace{-4mm}
\subsection{Preliminary baseline}

\paragraph{Diffusion for Capturing Invariant Features with $D(x^l;\xi)$ } 

We follow Diff-UNet~\cite{xing2023diffunet} and A\&D~\cite{wang2024towards} to use diffusion model to extract domain-invariant features. First, we  convert the label $y$ to the one-hot format $y_0\in\mathbb{R}^{Y\times L\times W\times H}$. During the forward process, successive $t$ steps of Gaussian noise are added 
to $y_0$, progressively transforming $y_0$ into a noisy version, as shown below:

\begin{equation}
\small
    \label{diffusion_forward}
    y_t=\sqrt{\bar{\alpha}_t}y_0+\sqrt{1-\bar{\alpha}_t}\epsilon, \epsilon \in \mathcal{N}(0,1)
\end{equation}
  Then, the diffusion encoder takes concatenation \verb|concat|$([y_t,x^l])$ and time step $t$ as input to generate the time-step-embedded multi-scale features $h^{l;\xi}_i \in \mathbb{R}^{i\times F \times \frac{D}{2^i}\times \frac{W}{2^i}\times \frac{H}{2^i}}$ where $i$ is the stage and $F$ is the basic feature size. $h^{l;\xi}_i$ is further used by $D(x^l;\xi)$ to predict the clear label $y_0$. The objective function is defined as follows:
\begin{equation}
\small
\label{eq:deno}
    \mathcal{L}_{diff}=\frac{1}{N_L}\sum_{i=0}^{N_L}\mathcal{L}_{DiceCE}(p^{l;\xi}_i, y_i),
\end{equation}
where $\mathcal{L}_{DiceCE}(x,y) = \frac{1}{2}[\mathcal{L}_{Dice}(x,y)+\mathcal{L}_{CE}(x,y)]$ is the combined dice and cross-entropy loss.

\paragraph{Supervised Training with $D(x^l;\psi)$}

  For the difficulty-aware training flow, \textit{i.e.}, $D(x^l;\psi)$  as decoder, the encoder only takes $x_l$ input to obtain the multi-scale features $h^{l;\psi}_i$. Note that $h^{l;\psi}_i$ are with the same shapes as $h^{l;\xi}_i$. The objective function of the supervised difficulty-aware training is defined as follows:
  \begin{equation}
\small
\label{eq:diff}
    \mathcal{L}_{sup}=\frac{1}{N_L}\frac{1}{K}\sum_{i=0}^{N_L}\sum_{k=0}^{K}  \mathcal{L}_{DiceCE}(p^{l;\psi}_{i,k}, y_{i,k})
\end{equation}
\subsection{Pseudo Labeling with Diverse Teaching  }

For the unlabeled data flow, \textit{i.e.},  $\mathcal{D}(x^u;\theta)$ as decoder, only takes  $x_u$ as input to obtain the multi-scale features $h^u_i$. To further train the $\mathcal{D}(x^u;\theta)$ branch, we design Diverse Teaching strategy to generate high quality pseudo labels, as illustrated in Fig.~\ref{fig:main} (b). 

\paragraph{Pseudo Labeling with Reparameterize \& Smooth (RS) Strategy}
The domain-unbiased $p^{u;\xi}$ probability map is generated by iterating the diffusion model ($\mathcal{E}(x^l, x^u;\xi)$+$\mathcal{D}(x^l;\xi)$) $t$ times with the Denoising Diffusion Implicit Models (DDIM) method~\cite{song2020ddim_diff}.
The class-unbiased $p^{u;\psi}$ probability map can be obtained by $\mathcal{D}(x^l;\psi)$ with stopped gradient forward pass.
We ensemble $p^{u;\xi}$ and $p^{u;\psi}$ to generate high-quality pseudo labels.
However, when combining these two maps, we found that the denoised probability map $p^{u;\psi}$ is too sparse, \textit{i.e.}, with very high confidence of each class.
This property is beneficial for fully-supervised tasks, but in this situation, it will suppress $p^{u;\psi}$ and is not robust to noise and error.
Thus, we re-parameterize $p^{u;\psi}$ with the Gumbel-Softmax to add some randomness and using Gaussian blur kernel to remove the noise brought by this operation.
% The final pseudo label is:
The final prediction is:
\begin{equation}
\small
\label{RS}
    p^{t1}= 0.5 * (\verb|Gumbel-Sofmax|(p^{u;\xi})+\verb|Softmax|(p^{u;\psi}))
\end{equation}

\paragraph{Pseudo Labeling with Mean Teacher model}

However, the Pseudo label generated via $p^{\xi, \psi}$ still contain noise, leading to suboptimal performance. Then, for  $\mathcal{D}(x^u;\theta)$ Decoder branch, we further obtain the mean-teacher Encoder $\mathcal{E}(x^l,x^u;\xi^{T})$ and  Decoder $\mathcal{D}(x^u;\theta^{T})$, where $\xi^{T} = \gamma \xi^{T} + (1-\gamma) \xi$, $\theta^{T} = \gamma \theta^{T} + (1-\gamma) \theta$. Given the unlabeled data $x^u$, the probability map can be obtained by $p^{t2}= \verb|Softmax|(p^{u;\theta^{T}})$

\paragraph{Unsupervised Training with $\mathcal{D}(x^u;\theta)$}

Given two distinct teacher models derived from $\mathcal{D}(x^l;\psi)$, $\mathcal{D}(x^u;\xi)$ and $\mathcal{D}(x^u;\theta^{T})$, we proceed to apply the entropy-based teacher ensemble method. This is done to obtain an ensembled prediction, $p^t$, for the unlabeled instance $u_i$. This prediction is then directly utilized for consistent supervision. Considering the prediction entropy (denoted as $H(\cdot)$) of the two teacher models,
\vspace{-2mm}
\begin{equation}
\small
    H_{t_1}  = -\sum_{i=1}^K p_{i}^{t_1} \log_2 p_{i}^{t_1}, 
    H_{t_2} =-\sum_{i=1}^K p_{i}^{t_2} \log_2 p_{i}^{t_2},
    % H_{t_1} &= H(q_{i}^{t_1}) = -\sum_{i=1}^K p_{i}^{t_1} \log_2 p_{i}^{t_1}, 
    % H_{t_2} &= H(q_{i}^{t_2}) =-\sum_{i=1}^K p_{i}^{t_2} \log_2 p_{i}^{t_2},
\end{equation}
we can obtain the entropy-based ensembled prediction,
\begin{align}
\label{eq:pseudo_label}
p^t_i &= \psi(q_{i}^{t_1}, q_{i}^{t_2}) = \frac{w_1 q_{i}^{t_1} + w_2 q_{i}^{t_2}}{w_1 + w_2},
\end{align}
with $w_1 = e^{-H_{t_1}}, w_2 = e^{-H_{t_2}}$. 

Finally, we can use the pseudo label $y^t=\verb|argmax|(p^t_i)$ to train $\mathcal{D}(x^u;\theta)$ in an unsupervised manner.
The objective function of the unsupervised training is defined as:
\begin{equation}
\small
\label{L_u}
    \mathcal{L}_{u}=\frac{1}{N_U}\sum_{i=0}^{N_U} \mathcal{L}_{DiceCE}(p^{u;\theta}_{i}, y^{\xi,\psi})
\end{equation}

% \vspace{-5mm}
\subsection{Global-local consistency learning}
To fully explore the data structure, we further adopt the cross-set CutMix strategy \cite{yun2019cutmix} and masked image modeling~\cite{he2022masked} to achieve global-local consistency learning, thus improving the model generalization and reducing domain shifts.

\paragraph{Global consistency learning with Cross-set CutMix} 

To further alleviate domain shifts, we expand the CutMix technique~\cite{yun2019cutmix} to cover both labeled and unlabeled datasets. This is accomplished by randomly generating a 3D binary mask $M \in \{0,1\}^{L\times H \times W}$ for a pair of volumes with:
\begin{equation}
\small
	\begin{aligned}
	x_{mix} = (\textbf{1} - M) \odot x_i + M  \odot x_j, \\
	y_{mix}^{t} = (\textbf{1} - M) \odot y_i^t + M  \odot y_j^t,
	\end{aligned}
 \label{eq:mix}
\end{equation}
where  $x_i \in D^{u}$ and  $x_j \in  D^{l}\cup  D^{u}$, $a\neq b$, $\textbf{1}\in\{1\}^{L\times W\times H}$, and $\odot$ means element-wise multiplication. Then, the objective function of the global consistency learning  is defined as:
\begin{equation}
\small
\label{L_mix}
    \mathcal{L}_{mix}=\frac{1}{N_U}\sum_{i=0}^{N_U} \mathcal{L}_{DiceCE}(p^{u;\theta}_{i,mix}, y^{t}_{i,mix}),
\end{equation}
where $p^{u;\theta}_{i,mix}= \mathcal{D} (\mathcal{E}(x_{i,mix}^{u}; \xi); \theta)$ .

\paragraph{Local Consistency Learning with masked image modeling }

Cross-Set CutMix consistency regularization focuses on learning the pairwise structure of the entire dataset. Nevertheless, when dealing with local regions within an individual image, it frequently encounters difficulties in attaining satisfactory segmentation outcomes. Consequently, we advocate specifically fostering the learning of context relations in unlabeled data. This approach is designed to offer supplementary cues, facilitating the robust recognition of classes that exhibit similar local appearances. Building in part upon this understanding, we have developed a Local Consistency Regularization (LCR).  This method is enabled by an auxiliary masked modeling proxy task, which serves to promote fine-grained locality learning. In addition, masked modeling consistency~\cite{hoyer2022mic} can also help in reducing the domain shifts. To this end, given a patch mask $\mathcal{M}$ that is randomly sampled from a uniform distribution:
\begin{equation}
\label{eq:mask}
    \mathcal{M}_{\substack{mb+1:(m+1)b,\\nb+1:(n+1)b}} = [v > r] \quad \text{with} \; v \sim \mathcal{U}(0, 1)\;,
\end{equation}
where $[\cdot]$ denotes the Iverson bracket, $b$ the patch size, $r$ the mask ratio, and $m\in[0 \dots W/b-1]$, $n\in[0 \dots W/b-1]$ the patch indices.
The masked target image  $x_{i,mask} = \mathcal{M} \odot x_i$  is obtained by element-wise multiplication of mask and image. Then the local consistency learning with masked image modeling can be written as:
\begin{equation}
\label{eq:mic}
    \mathcal{L}_{mic}=\frac{1}{N_U}\sum_{i=0}^{N_U} \mathcal{L}_{DiceCE}(p^{u;\theta}_{i,mask}, y^{t}_{i}),
\end{equation}
where $p^{l;\theta}_{i,mask}= \mathcal{D} (\mathcal{E}(x_{i,mask}^{u}; \xi); \theta)$ .

\vspace{-3mm}
\subsection{Knowledge Distillation and Masked  Reconstruction}
The hard pseudo labels generated by the dual teachers via eq.~\ref{eq:pseudo_label} still inevitably contain noise, especially
in the scenario where there is class imbalance. Therefore, for the unlabeled data, we distill knowledge from the Decoder branches $\mathcal{D}(x^l;\psi)$ and $\mathcal{D}(x^l,x^u;\xi)$ to the student $\mathcal{D}(x^u;\theta)$ decoder, respectively. This is accomplished by minimizing the soft dice loss $\mathcal{L}_{sdice}$~\cite{milletari2016v} using the predictions of $p^{u;\xi}$ and $p^{u;\psi}$ . 
\begin{equation}
\small
\label{eq:kd}
    \mathcal{L}_{kd} = \frac{1}{N_U}\sum_{i=0}^{N_U} \left(\mathcal{L}_{sdice}\left(p^{u;\theta}_{i}, p^{u;\xi}\right) + \mathcal{L}_{sdice}\left(p^{u;\theta}_{i}, p^{u;\psi}\right)\right).
\end{equation}
  
In addition, we further try to exploit the soft predictions generated by the mean-teacher Encoder $\mathcal{E}(x^l,x^u;\xi^{T})$ and  Decoder $\mathcal{D}(x^u;\theta^{T})$ to reconstruct the masked student, thereby capturing the contextual of the medical image with less noise. Specifically, we further  minimize the difference between the voxel-wise logits generated by the second teacher network  given the original input $\bm{x}_{i}$ and the voxel-wise logits produced by the student   using the masked input $\bm{x}_{i,mask}$. The objective function is expressed as,
\begin{equation}
\small
\label{eq:rec}
    \mathcal{L}_{rec} = \frac{1}{N_L+N_U}\sum_{i=0}^{N_L+N_U} \frac{\Vert\ p^{u,l;\theta}_{i,mask} - p^{u,l;\theta^{t}}_{i}  \Vert\ _{2}^{2}}{\Vert\ p^{u,l;\theta^{t}}_{i}  \Vert\ _{2}^{2}}.
\end{equation}

% To furthermore reduce the potential noise in hard  pseudo labels generated  via eq.~\ref{eq:pseudo_label}, for the unlabeled data, we distill knowledge from the Decoder branches $\mathcal{D}(x^l;\psi)$ and $\mathcal{D}(x^l,x^u;\xi)$ to the $\mathcal{D}(x^u;\theta)$ decoder, respectively. This is accomplished by minimizing the soft dice loss $\mathcal{L}_{sdice}$~\cite{milletari2016v} using the predictions of $p^{u;\xi}$ and $p^{u;\psi}$ . 
% \begin{equation}
% \small
% \label{eq:kd}
%     \mathcal{L}_{kd} = \frac{1}{N_U}\sum_{i=0}^{N_U} \left(\mathcal{L}_{sdice}\left(p^{u;\theta}_{i}, p^{u;\xi}\right) + \mathcal{L}_{sdice}\left(p^{u;\theta}_{i}, p^{u;\psi}\right)\right).
% \end{equation}

\vspace{-4mm}
% \subsection{voxel-level correlation}
\subsection{Voxel-level Label Propagation} \label{sec:correlation-maps}

The global-local consistency learning can  explore the inter-sample and intra-sample data structure. However, the pairwise similarities on the voxel-level remain to be solved. In this section, we propose the voxel-level propagation strategy to fully explore the potential of unlabeled data. First, we extract features $e_1$ and $e_2 \in \mathbb{R}^{D \times LHW}$ through linear layers after the encoder of the network, where $D$ is the channel dimension and $LHW$ is the number of feature vectors.

These extracted features enable correlation matching to quantify the degree of pairwise similarity.
% given the extracted features $w_1$ and $w_2$, 
Thus, we compute the correlation map $\mathcal{C}$ by performing a matrix multiplication between all pairs of feature vectors:
\begin{equation}
    \mathcal{C} = \mathrm{Softmax}({e}_1^{\top}\cdot {e}_2)/\sqrt{D},
\end{equation}
where $^\top$ denotes the matrix transpose operation. 
The correlation map $\mathcal{C} \in \mathbb{R}^{LHW \times LHW}$ is a 3D matrix and is activated by a $\mathrm{Softmax}$ function to yield pairwise similarities. 
$\mathcal{C}$ enables accurate delineation of the corresponding regions belonging to the same object and inspires us to propagate it into pseudo labels using correlation matching. 
% More visualizations can be found in \figref{fig:enhance}.

To enhance the model's awareness of pairwise similarity, we spread the correlation map $\mathcal{C}$ into model logits outputs $\mathcal{D}(x^{u}_{i}; \theta)$ to attain another representation of the prediction $\mathbf{z}_i^u \in  \mathbb{R}^{ K \times LHW}$ via label propagation:
\begin{equation}
\label{eqn:reaction}
    p_{i,corr}^{u;\theta} = \mathcal{D}(\mathcal{E}(x^{u}_{i}; \xi); \theta) \cdot  \mathcal{C},
\end{equation}
% where $f_1(\cdot)$ is a bilinear interpolation for shape matching. 
%
the resulting $p_{i,corr}^{u;\theta}$ emphasizes the pairwise similarities of the same object through the correlation map. Then, a correlation loss $\mathcal{L}_{corr}^{u}$ can be calculated as follows:
\begin{equation}
\label{eqn:correlation loss}
    \mathcal{L}_{corr}^{u} = \frac{1}{N_U} \sum^{N_U}_{i=1} (\mathcal{L}_{DiceCE}( p_{i,corr}^{u;\theta}, y_i^t)). %\odot \mathcal{M}_i.
\end{equation}
Similarly, for the labeled images branch $\mathcal{D}(x_i^{l}; \xi)$, we also compute the $\mathcal{L}_{corr}^{l}$. Then, the correlation loss is obtained via:
\begin{equation}
\label{eq:corr}
    \mathcal{L}_{corr} =  \mathcal{L}_{corr}^{u} +  \mathcal{L}_{corr}^{l}.%\odot \mathcal{M}_i.
\end{equation}

\vspace{-4mm}
\subsection{Total Taring loss}

Integrating these objectives introduced in Eqs.~\ref{eq:deno}, \ref{eq:diff}, \ref{eq:mix}, ~\ref{eq:mic}, ~\ref{eq:kd}, ~\ref{eq:rec}, ~\ref{eq:corr} together, the final loss function as follows:

\begin{equation} %\label{eq_total}
\small
\label{eq:total}
   \mathcal{L} = \mathcal{L}_{diff} + \mathcal{L}_{sup} + \mathcal{L}_{mix}  + \alpha  \mathcal{L}_{mic} + \beta  \mathcal{L}_{kd} + \gamma \mathcal{L}_{rec}  + \eta  \mathcal{L}_{corr},
\end{equation}
where $\alpha$, $\beta$, $\gamma$, and $\eta$ are the parameters to control the importance of the loss.

\section{Experiments}
\label{sec.exp}

\subsection{Datasets and Implementation Details}

Our method is evaluated on five publicly accessible datasets, including three semi-supervised benchmark datasets, \textit{i.e.}, the LA dataset~\cite{xiong2021laseg}, Synapse dataset~\cite{synapse},  and AMOS dataset~\cite{ji2022amos}, one dataset MMWHS~\cite{zhuang2016mmwhs} for UMDA, and one dataset M\&Ms~\cite{prados2017spinal} for Semi-MDG. To assess the network's prediction, the Dice metric and the average surface distance (ASD) are employed. In the context of SSMIS tasks, following previous works~\cite{li2024diversity}, the additional Jaccard and HD95 metrics are also utilized.

All of our experiments were implemented with PyTorch 1.12.1 on one NVIDIA A100 GPU. For the shared encoder $\mathcal{E}(x^l,x^u;\xi)$ and Diffusion Decoder  $\mathcal{D}(x^l,x^u;\xi)$, We follow Diff-UNet~\cite{xing2023diffunet} and A\&D~\cite{wang2024towards} to adopt  diffusion model for perception but modify it to a V-Net version and remove the additional image encode. For the Decoders  $\mathcal{D}(x^l;\phi)$ and  $\mathcal{D}(x^l;\theta)$, a V-Net Decoder~\cite{milletari2016v} is adopted. We employ the Stochastic Gradient Descent (SGD) optimizer and utilize polynomial scheduling to adapt the learning rate. The learning rate is adjusted using the formula  $lr = lr_{init} \cdot (1 - \frac{i} {I} )^{0.9} $, where $lr_{init}$ denotes the initial learning rate, $i$ is the current iteration, and $I$ is the maximum number of iterations. The training epoch is set to 300. The batch size is set to 4, with 2 labeled and 2 unlabeled data. The mask ratio in Eq.~\ref{eq:mask} is set to 0.5 except for 0.7 for the LA dataset, and the block size is set to 1/16 of the image size.
The patch size, learning rate, and optimal values for $\alpha$, $\beta$ , $\gamma$, and $\eta$ in Eq.~\ref{eq:total}  are summarized  in Tab.~\ref{tab:parameters}.

\begin{table}[!hb]\footnotesize
	\centering

 \caption{Parameter settings for different datasets.}
 \vspace{-2mm}
	\setlength{\tabcolsep}{2pt}
	\begin{tabular}{l|ccccccc}
	\hline
	Parameters & patch size   & learning rate  &  mask ratio $r$ & $\alpha$ & $\beta$   & $\gamma$ & $\eta$   \\
	\hline
         LA   & $112\times112\times80$  & 1e-2 & 0.7 & 2.0 & 0.1 & 0.2 & 1.2   \\
         Synapse & $64\times128\times128$  & 3e-2 & 0.5 & 0.1 & 0.2 & 0.5 & 1.0    \\
         AMOS  & $64\times128\times128$  & 3e-2 & 0.5 & 0.1 & 0.2 & 0.5 & 1.3     \\
         MMWHS  & $128\times128\times128$     &  5e-3   & 0.5 & 1.0 & 0.1 & 0.1 & 0.9    \\
         M\&Ms & $32\times128\times128$      &   1e-2  & 0.5 & 1.0 & 0.1 & 0.1 & 1.0    \\
	\hline
	\end{tabular}
	\label{tab:parameters}
\end{table}

\vspace{-4mm}
\subsection{Competitors}
% \vspace{-2mm}
We compare our proposed method with SSMIS, UMDA and Semi-MDG methods, including:

% \vspace{-2mm}
\paragraph{SSMIS methods}  \textbf{General SSMIS methods} : UA-MT~\cite{yu2019uamt}; URPC~\cite{luo2021urpc}; CPS~\cite{chen2021cps};DTC \cite{luo2021dtc}; SASSNet \cite{li2020sassnet}; MC-Net~\cite{wu2021mcnet}; SS-Ne~\cite{wu2022exploring};  DePL~\cite{wang2022depl}; BCP~\cite{bai2023bidirectional}; OTCMC~\cite{guo2025optimal}; GapMatch~\cite{huang2025gapmatch};  w2sPC~\cite{yang2025semi}. \textbf{Imbalance SSMIS methods}: Adsh~\cite{guo2022adsh} ; SimiS~\cite{simis}; DHC~\cite{wang2023dhc}; AllSpark~\cite{wang2024allspark} ; A\&D~\cite{wang2024towards}; InterTeach~\cite{zhang2024interteach}; SKCDF~\cite{zhang2025semantic}.

% \vspace{-2mm}
\paragraph{UMDA methods} PnP-AdaNet~\cite{dou2019pnp_uda}, CycleGAN~\cite{zhu2017cyclegan_uda}, CyCADA~\cite{hoffman2018cycada_uda} , DSAN~\cite{han2021dsan_uda}, LMISA-3D~\cite{jafari2022lmisa_uda}, DDSP~\cite{zheng2024dual}.
% ;AdaOutput~\cite{tsai2018adaouput_uda} SIFA~\cite{chen2020sifa_uda}, 
% \vspace{-2mm} 
\paragraph{Semi-MDA methods} SDNet+Aug~\cite{liu2021SDNet} , LDDG~\cite{li2020lddg}, SAML~\cite{liu2020saml}, DGNet~\cite{liu2021dgnet} , vMFNet~\cite{liu2022vmfnet}, Meta~\cite{liu2021semi},  StyleMatch~\cite{zhou2023semi}, EPL~\cite{yao2022enhancing}.

% LDDG~\cite{li2020lddg}, ,  SAML~\cite{liu2020saml}

\begin{table}[t] \small 
\caption{Comparisons on the LA dataset with 5\% and 10\% labeled data, respectively.}
\vspace{-2mm}
\setlength{\tabcolsep}{0.9mm}{
\resizebox{1\linewidth}{!}{
\begin{tabular}{c|cc|llll}
\toprule
\multicolumn{1}{c|}{\multirow{2}{*}{Method}} & \multicolumn{2}{c|}{Scans used}  & \multicolumn{4}{c}{Metrics} \\ 
\cline{2-7} \multicolumn{1}{c|}{} & \multicolumn{1}{l}{Labeled} & \multicolumn{1}{l|}{Unlabeled} & 
Dice$\uparrow$ & Jaccard$\uparrow$ & 95HD$\downarrow$ & ASD$\downarrow$ \\ \hline
\multicolumn{1}{c|}{V-Net} &\multicolumn{1}{c}{4(5\%)} &\multicolumn{1}{c|}{0} &52.55 &39.60 &47.05 &9.87 \\ 
V-Net (fully) &\multicolumn{1}{c}{80(100\%)} &\multicolumn{1}{c|}{0}  & 91.47  &  84.36   & 5.48     & 1.51 \\
\hline
UA-MT \cite{yu2019uamt}  & \multirow{11}{*}{4(5\%)} & \multirow{11}{*}{76(95\%)} 
& 82.26 & 70.98 & 13.71 & 3.82 \\
URPC \cite{luo2021urpc}  & & & 82.48 & 71.35 & 14.65 & 3.65 \\
DTC \cite{luo2021dtc}  & & & 81.25 & 69.33 & 14.90 & 3.99 \\
SASSNet \cite{li2020sassnet} &  &  & 81.60 & 69.63 & 16.16 & 3.58 \\ 
MC-Net \cite{wu2021mcnet}  & & & 83.59 & 72.36 & 14.07 & 2.70  \\
SS-Net \cite{wu2022exploring}  & & & 86.33 & 76.15 & 9.97 & 2.31 \\ 
BCP \cite{bai2023bidirectional}  & & & 88.02 & 78.72 & 7.90 & 2.15 \\ 
AllSpark~\cite{wang2024allspark} & & & 87.99   & 78.83    &  7.44    & 2.10   \\
A\&D~\cite{wang2024towards}  & & &  \color{blue}\textbf{89.93}   & \color{blue}\textbf{81.82}    & \color{blue}\textbf{5.25}    & {1.86}   \\ 
InterTeach~\cite{zhang2024interteach} & & & {89.76}   & \color{blue}\textbf{81.49} & \color{blue}\textbf{6.31} & {1.75}   \\
w2sPC~\cite{yang2025semi} & & & 89.02  & 79.83 & 10.23 & 2.18 \\
GapMatch~\cite{huang2025gapmatch} &                            &           &  88.3   & -   & - & \color{blue}\textbf{1.77} \\
OTCMC\cite{guo2025optimal}  & & & 87.97  & 78.65  & 8.39 & 2.02 \\
\rowcolor{red!10} Ours & & & \color{red}\textbf{91.63}{\color{right}\textbf{\scriptsize{$\uparrow$}1.70}} &\color{red}\textbf{84.59}{\color{right}\textbf{\scriptsize{$\uparrow$}2.77}} &\color{red}\textbf{4.40}{\color{right}\textbf{\scriptsize{$\downarrow$}0.85}} &\color{red}\textbf{1.41}{\color{right}\textbf{\scriptsize{$\downarrow$}0.36}} \\                 
\bottomrule
\bottomrule
% \multicolumn{1}{c|}{\multirow{2}{*}{Method}} & \multicolumn{2}{c|}{Scans used}  & \multicolumn{4}{c}{Metrics} \\ 
% \cline{2-7} \multicolumn{1}{c|}{} & \multicolumn{1}{l}{Labeled} & \multicolumn{1}{l|}{Unlabeled} & 
% Dice$\uparrow$ & Jaccard$\uparrow$ & 95HD$\downarrow$ & ASD$\downarrow$ \\ \hline
\multicolumn{1}{c|}{V-Net} &\multicolumn{1}{c}{8(10\%)} &\multicolumn{1}{c|}{0} &82.74 &71.72 &13.35 &3.26 \\ 
\hline
UA-MT \cite{yu2019uamt}  & \multirow{11}{*}{8(10\%)} & \multirow{11}{*}{72(90\%)} 
& 87.79 & 78.39 & 8.68 & 2.12 \\
URPC \cite{luo2021urpc}  & & & 86.92 & 77.03 & 11.13 & 2.28  \\ 
DTC \cite{luo2021dtc}  & & & 87.51 & 78.17 & 8.23 & 2.36  \\
SASSNet \cite{li2020sassnet}  & & & 87.54 & 78.05 & 9.84 & 2.59 \\ 
MC-Net \cite{wu2021mcnet} & & & 87.62 & 78.25 & 10.03 & 1.82   \\
SS-Net \cite{wu2022exploring}  & & & 88.55 & 79.62 & 7.49 & 1.90 \\
% Simcvd \cite{you2022simcvd} & & & 89.03 & 80.34 & 8.34 & 2.59 \\
BCP \cite{bai2023bidirectional}  & & & 89.62 & 81.31 & 6.81 & 1.76 \\
% ACTION++ \cite{you2023action++}  & & & 89.9 & NA & NA & 1.74   \\ 
AllSpark~\cite{wang2024allspark} & &  & 88.74  &80.54 & 7.06 & 1.82   \\
% MLRPL \cite{su2024mutual}  & & & 89.86 & 81.68 & 6.91 & 1.85 \\
A\&D~\cite{wang2024towards} & & & {90.31}   & \color{blue}\textbf{82.40} & \color{blue}\textbf{5.55} & {1.64}   \\
InterTeach~\cite{zhang2024interteach} & & & {91.11}   & \color{blue}\textbf{83.72} & \color{blue}\textbf{5.19} & {1.55}   \\
% CMMT-Net~\cite{li2024diversity} & & &\color{blue}{90.75}   &\color{blue}{83.13} &\color{blue}{5.77} &\color{blue}{1.82}   \\ 
w2sPC~\cite{yang2025semi}  & & & 90.23 & 81.52 & 7.16 & 1.95 \\ % 
GapMatch~\cite{huang2025gapmatch}  &  &  &  \color{blue}\textbf{91.0}   & -   & - & \color{blue}\textbf{1.46} \\
OTCMC\cite{guo2025optimal} & & & 90.26 & 82.34 & 5.96 & 1.65 \\

\rowcolor{red!10} Ours & & & \color{red}\textbf{92.43}{\color{right}\textbf{\scriptsize{$\uparrow$}1.43}} 
&\color{red}\textbf{85.96}{\color{right}\textbf{\scriptsize{$\uparrow$}3.56}}
&\color{red}\textbf{3.94}{\color{right}\textbf{\scriptsize{$\downarrow$}1.61}}
&\color{red}\textbf{1.30}{\color{right}\textbf{\scriptsize{$\downarrow$}0.16}}\\ 
\bottomrule    
\end{tabular}}
}
\label{tab:LA_4}
\end{table}

\begin{table*}[t]
\centering
\scriptsize
\caption{
Results on Synapse dataset with 20\% labeled data for \textbf{class imbalanced SSL} task. `Common' or `Imbalance' indicates whether the methods consider the imbalance issue or not.  Results of 3-times repeated experiments are reported in `mean$\pm$std' format.  %Notably, our method does not introduce many additional parameters compared with the existing SSL methods.
}
\vspace{-2mm}
\label{tab:Sy_20}
\setlength\tabcolsep{2pt}
\resizebox*{1.0\linewidth}{!}{

\begin{tabular}{c|c|@{\quad }cc@{\quad }|ccccccccccccc}
\toprule

\multicolumn{2}{c|@{\quad }}{\multirow{2}{*}{Methods}}  & {Avg.} & {Avg.} & \multicolumn{13}{c}{Dice of Each Class}            \\ 
\multicolumn{2}{c|@{\quad }}{}                          &   Dice~$\uparrow$                         &   ASD~$\downarrow$                       & Sp   & RK   & LK   & Ga   & Es   & Li   & St   & Ao   & IVC  & PSV  & PA   & RAG  & LAG  \\\midrule
\multirow{6}{*}{\rotatebox{90}{General}}         & V-Net (fully)      & 62.09$\pm$1.2	&10.28$\pm$3.9	& 84.6	& 77.2	& 73.8	& 73.3	& 38.2	& 94.6	& 68.4	& 72.1	& 71.2	& 58.2	& 48.5	& 17.9	& 29.0 \\ \midrule
& UA-MT~\cite{yu2019uamt}   & 20.26$\pm$2.2	&71.67$\pm$7.4	& 48.2	& 31.7	& 22.2	& 0.0	& 0.0	& 81.2	& 29.1	& 23.3	& 27.5	& 0.0	& 0.0	& 0.0	& 0.0  \\
% & URPC~\cite{luo2021urpc}      & 25.68$\pm$5.1	&72.74$\pm$15.5	& 66.7	& 38.2	& 56.8	& 0.0	& 0.0	& 85.3	& 33.9	& 33.1	& 14.8	& 0.0	& 5.1	& 0.0	& 0.0  \\
% & CPS~\cite{chen2021cps}     & 33.55$\pm$3.7	&41.21$\pm$9.1	& 62.8	& 55.2	& 45.4	& 35.9	& 0.0	&\textbf{91.1}	& 31.3	& 41.9	& 49.2	& 8.8	& 14.5	& 0.0	& 0.0  \\
& SS-Net~\cite{wu2022exploring}  & 35.08$\pm$2.8	&50.81$\pm$6.5	& 62.7	& 67.9	& 60.9	& 34.3	& 0.0	& 89.9	& 20.9	& 61.7	& 44.8	& 0.0	& 8.7	& 4.2	& 0.0  \\
% & DST~\cite{chen2022dst}      & 34.47$\pm$1.6	&37.69$\pm$2.9	& 57.7	& 57.2	& 46.4	& 43.7	& 0.0	& 89.0	& 33.9	& 43.3	& 46.9	& 9.0	& {21.0}	& 0.0	& 0.0  \\
& DePL~\cite{wang2022depl}       & 36.27$\pm$0.9	&36.02$\pm$0.8	& 62.8	& 61.0	& 48.2	& 54.8	& 0.0	& {90.2}	& {36.0}	& 42.5	& 48.2	& 10.7	& 17.0	& 0.0	& 0.0  \\ \midrule
\multirow{6}{*}{\rotatebox{90}{Imbalance}} 
& Adsh~\cite{guo2022adsh}      & 35.29$\pm$0.5	&39.61$\pm$4.6	& 55.1	& 59.6	& 45.8	& 52.2	& 0.0	& 89.4	& 32.8	& 47.6	& 53.0	& 8.9	& 14.4	& 0.0	& 0.0  \\ 
% & CReST~\cite{wei2021crest}      & 38.33$\pm$3.4	&{22.85$\7pm$9.0}	& 62.1	& 64.7	& 53.8	& 43.8	& {8.1}	& 85.9	& 27.2	& 54.4	& 47.7	& 14.4	& 13.0	& {18.7}	& 4.6  \\
& SimiS~\cite{simis}      & 40.07$\pm$0.6	&32.98$\pm$0.5	& 62.3	& {69.4}	& 50.7	& 61.4	& 0.0	& 87.0	& 33.0	& 59.0	& {57.2}	& {29.2}	& 11.8	& 0.0	& 0.0  \\
% & Basak \textit{et al.}~\cite{basak2022addressing}       &33.24$\pm$0.6	&43.78$\pm$2.5	& 57.4	& 53.8	& 48.5	& 46.9	& 0.0	& 87.8	& 28.7	& 42.3	& 45.4	& 6.3	& 15.0	& 0.0	& 0.0   \\          
% & CLD~\cite{lin2022cld}      &{41.07$\pm$1.2}	&32.15$\pm$3.3	& 62.0	& 66.0	& {59.3}	& \color{blue}\textbf{61.5}	& {0.0}	& 89.0	& 31.7	& {62.8}	& 49.4	& 28.6	& 18.5	& {0.0}	& {5.0}  \\
& DHC~\cite{wang2023dhc}   & 48.61$\pm$0.9	&10.71$\pm$2.6	& 62.8	& 69.5	& 59.2	& \color{red}\textbf{66.0}	& 13.2	& 85.2	& 36.9	& 67.9	& 61.5	& 37.0	& 30.9	& 31.4	& 10.6 \\ 
 & A\&D~\cite{wang2024towards}    & 60.88$\pm$0.7	& 2.52$\pm$ 0.4	& 85.2 & 66.9 & 67.0 & 52.7  & \color{red}\textbf{62.9} &89.6 &   52.1 & 83.0 & 74.9 & 41.8 & 43.4 & 44.8 & 27.2 \\ 
 & AllSpark~\cite{wang2024allspark}   & 60.68 $\pm$ 0.6	& 2.37$\pm$ 0.3	& \color{blue}\textbf{86.3} & \color{red}\textbf{79.6} & \color{red}\textbf{77.8}  & 60.4  & 60.7 & \color{blue}\textbf{92.3} & \color{blue}\textbf{63.7} & 75.0 & 69.9 & \color{red}\textbf{60.2} & \color{red}\textbf{57.7} & 0.0 & 5.2 \\
 &  SKCDF~\cite{zhang2025semantic} & \color{blue}\textbf{64.27 $\pm$ 1.36} & \color{red}\textbf{1.45 $\pm$ 0.09} & 79.5 & 72.1 & 67.6 & 59.8 & 60.7 & 93.3 & 61.7 & \color{blue}\textbf{85.4} & \color{blue}\textbf{78.5} & 41.8 & 50.9 & \color{red}\textbf{46.4} & \color{red}\textbf{37.8} \\
\midrule
\rowcolor{red!10} & Ours &  \color{red}\textbf{66.01$\pm$0.43}{\color{right}\textbf{\scriptsize{$\uparrow$}1.74}}  & \color{blue}\textbf{1.69$\pm$0.79}{\color{right}\textbf{\scriptsize{$\uparrow$}0.24}}    & \color{red}\textbf{89.2}  & \color{blue}\textbf{75.0}  & \color{blue}\textbf{75.0}  & 51.4 & \color{blue}\textbf{62.3} & \color{red}\textbf{93.6} & \color{red}\textbf{64.8} & \color{red}\textbf{85.7} & \color{red}\textbf{78.6} & \color{blue}\textbf{47.8} & \color{blue}\textbf{52.6} & \color{blue}\textbf{45.3} & \color{blue}\textbf{36.8} \\ 
\bottomrule
\end{tabular} 
}
\end{table*}

\begin{table}[!ht]
% \footnotesize
\centering
\caption{Quantitative comparisons on \textbf{10\% labeled Synapse dataset.} }
\vspace{-2mm}
\label{tab:Sy_10}
\resizebox*{0.8\linewidth}{!}{
\begin{tabular}{c|c|cc}
\toprule
% \multicolumn{2}{c|}{\multirow{2}{*}{Methods}}  & \multirow{2}{*}{Avg. Dice~$\uparrow$} & \multirow{2}{*}{Avg. ASD~$\downarrow$}                                        \\ 
% \multicolumn{2}{c|}{}                          &                            &                           \\\midrule
\multicolumn{2}{c|}{Methods}  & Avg. Dice~$\uparrow$ &Avg. ASD~$\downarrow$                                      \\ \midrule
                         % &                            &                           \\\midrule
\multirow{6}{*}{\rotatebox{90}{Imbalance}} 

& Adsh~\cite{guo2022adsh}$^\star$       &22.8$\pm$0.9	&46.18$\pm$4.0	 \\ 

& SimiS~\cite{simis}$^\star$      & 25.05$\pm$3.1	&43.93$\pm$2.4	 \\

% & Basak \textit{et al.}~\cite{basak2022addressing}$^\dagger$         &25.3$\pm$2.2	&50.02$\pm$5.7	& 40.9	& 42.3	& 19.2	& 35.2	& \textcolor{red}{0.0}	& 75.7	& 19.2	& 44.7	& 32.8	& 5.0	& \textbf{10.4}	& 3.5	& \textcolor{red}{0.0}  \\
                                 
% & CLD~\cite{lin2022cld}$^\dagger$        & 22.49$\pm$1.6	&49.74$\pm$4.1	  \\

& DHC~\cite{wang2023dhc}    & 31.64$\pm$0.9	& 21.82$\pm$1.0	 \\   
 & A\&D~\cite{wang2024towards}    & 46.24$\pm$0.8	&\color{blue}\textbf{7.78$\pm$2.13}	 \\ 
& SKCDF~\cite{zhang2025semantic} & \color{blue}\textbf{48.45 $\pm$ 0.6} & \color{blue}\textbf{7.87 $\pm$ 3.47} \\
% & 73.5 & 66.7 & 64.2 & 24.4 & 27.2 & 90.0 & 30.3 & 79.0 & 68.5 & 33.7 & 18.1 & 37.3 & 17.1 \\

 \midrule
\rowcolor{red!10} & Ours &  \color{red}\textbf{59.43$\pm$0.99}{$\color{right}\textbf{\scriptsize{$\uparrow$}10.98}$}& 
 \color{red}\textbf{1.51$\pm$0.05}{\color{right}\textbf{\scriptsize{$\downarrow$}6.36}} \\ 
\bottomrule
\end{tabular}
}
\end{table}

\begin{table}[!htbp]
% \scriptsize
\caption{RESULTS ON AMOS DATASET FOR SSMIS TASK.}
% \vspace{-2mm}
    \label{tab:amos}
    \setlength\tabcolsep{3pt}
    \centering
    \resizebox*{0.5\textwidth}{!}{
    \begin{tabular}{lcccccc}
    \toprule
    \multicolumn{7}{c}{2\% labeled data (labeled:unlabeled=4:212)} \\ 
    \midrule
    Metrics & V-Net (fully) & CPS~\cite{chen2021cps} & DHC~\cite{wang2023dhc} &  $A\&D$~\cite{wang2024towards}  & AllSpark~\cite{wang2024allspark} & \cellcolor{red!10} \textbf{Ours}  \\ 
    \midrule
    Avg. Dice~$\uparrow$ & 76.50$\pm$2.32 & 31.78$\pm$5.44 & 38.28$\pm$1.93  &  36.05$\pm$1.41 & \color{blue}\textbf{40.20$\pm$2.29} &   \cellcolor{red!10}\color{red}\textbf{43.18$\pm$0.52}{\color{right}\textbf{\scriptsize{$\uparrow$}2.98}} \\
    Avg. ASD~$\downarrow$ & 2.01$\pm$1.47 & 39.23$\pm$7.24 & 20.34$\pm$4.22  &  18.60$\pm$0.61 & \color{blue}\textbf{14.77$\pm$2.88}  &  \cellcolor{red!10}\color{red}\textbf{13.30$\pm$1.46}{\color{right}\textbf{\scriptsize{$\downarrow$}1.77}} \\ 
    \bottomrule 
    \toprule
    \multicolumn{7}{c}{5\% labeled data (labeled:unlabeled=11:205)} \\
    \midrule
    Metrics & V-Net (fully) & CPS~\cite{chen2021cps} & DHC~\cite{wang2023dhc} &  $A\&D$~\cite{wang2024towards} & AllSpark~\cite{wang2024allspark} & \cellcolor{red!10}{\textbf{Ours}}  \\ 
    \midrule
    Avg. Dice~$\uparrow$ & 82.39$\pm$3.64 & 41.08$\pm$3.09 & 49.53$\pm$2.22   & 52.17$\pm$1.45  & \color{blue}\textbf{53.77$\pm$1.88} &  \cellcolor{red!10}\color{red}\textbf{55.13$\pm$0.49}{\color{right}\textbf{\scriptsize{$\uparrow$}1.36}} \\
    Avg. ASD~$\downarrow$ & 1.19$\pm$0.67 & 20.37$\pm$2.97 & 13.89$\pm$3.64  &  \color{red}\textbf{4.66$\pm$0.22}  & 10.96 $\pm$ 2.28  &  \cellcolor{red!10}\color{blue}\textbf{6.71$\pm$2.55}{\color{right}\textbf{\scriptsize{$\uparrow$}2.05}} \\ 
    \bottomrule 
    \end{tabular}
    }
    % \vspace{-0.7cm}
\end{table}

\begin{figure}[!htbp]
	\begin{center}
		\centering
		\includegraphics[width=.5\textwidth]{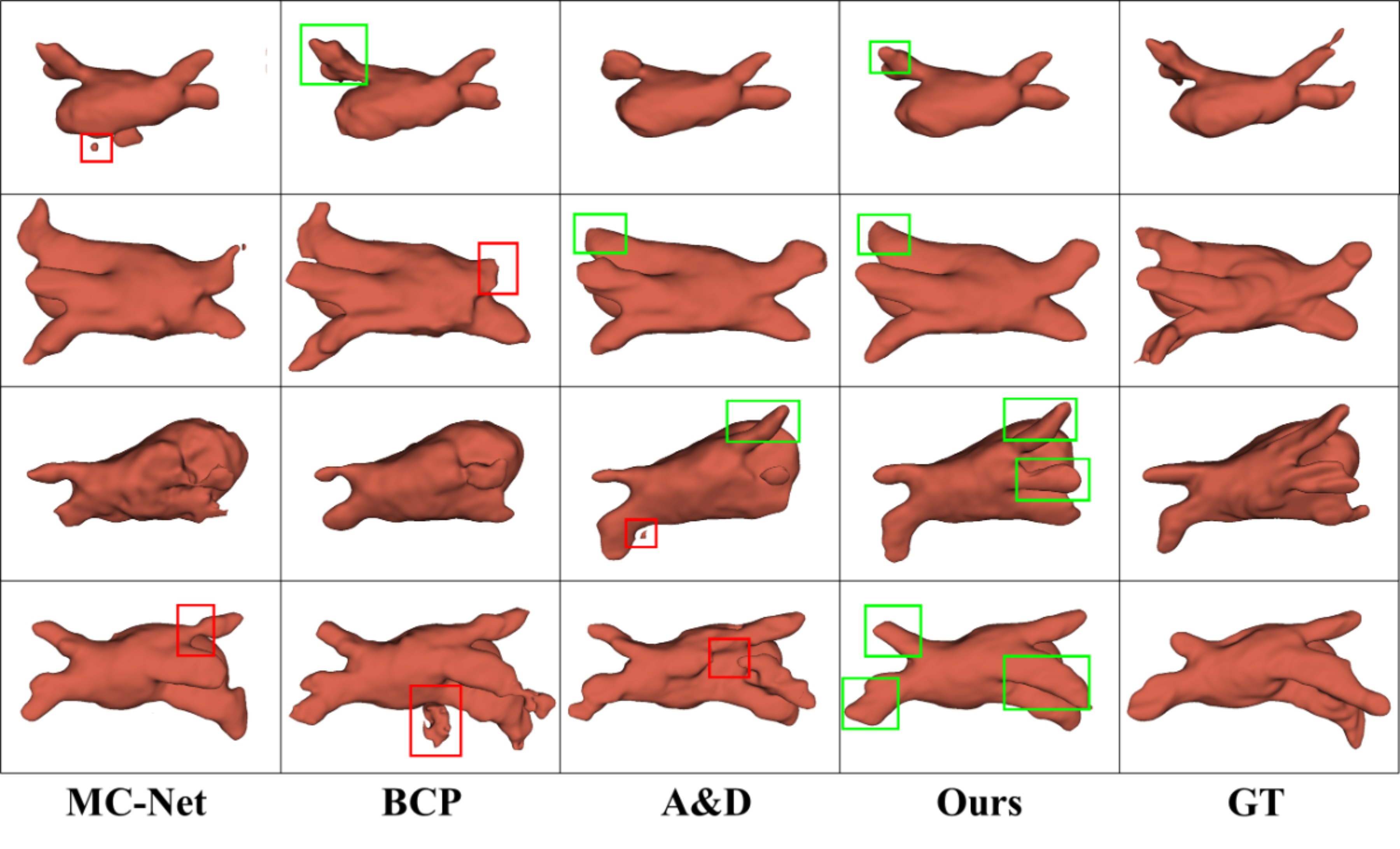}
		\centering
        % \vspace{-5mm}
		\caption{Visualization results with 5\% labeled data on the LA dataset.}
        % \vspace{-5mm}
		\label{fig:la}
	\end{center}
\end{figure}

\begin{table}[!htbp]
\caption{Results on two settings, \textit{i.e.}, MR to CT and CT to MR, of MMWHS dataset for \textbf{UMDA} task.}
% \vspace{-2mm}
\label{tab:UMDA}
\setlength\tabcolsep{3pt}
\centering
\resizebox*{0.45\textwidth}{!}{
\begin{tabular}{lcccccc}
\toprule
\rowcolor{cyan!20} \multicolumn{7}{c}{MR to CT} \\ \toprule
\multirow{2}{*}{Method}    & \multicolumn{5}{c}{Dice~$\uparrow$} & \multicolumn{1}{c}{ASD~$\downarrow$}\\
& AA  & LAC &   LVC & MYO & Average  & Average \\ \midrule
Vnet (Fully) &   92.7  &  91.1   & 91.9  & 87.8   &  90.9  &2.2 \\ \midrule
PnP-AdaNet~\cite{dou2019pnp_uda} &  74.0     & 68.9    & 61.9    & 50.8  &63.9    & 12.8  \\
% AdaOutput~\cite{tsai2018adaouput_uda}    &     65.2      &      76.6      &  54.4   &  43.6   & 59.9       &9.6   \\
CycleGAN~\cite{zhu2017cyclegan_uda}    &     73.8      &      75.7      &  52.3   &  28.7   &  57.6       &10.8   \\
CyCADA~\cite{hoffman2018cycada_uda}    &     72.9      &      77.0      &  62.4   &  45.3   &   64.4       &9.4  \\
% SIFA~\cite{chen2020sifa_uda}    &     81.3      &      79.5      &  73.8   &  61.6   &    74.1       &7.0\\
% DSFN~\cite{zou2020dsfn_uda}    &     84.7      &      76.9      &  79.1   &  62.4   &    75.8       &- \\
DSAN~\cite{han2021dsan_uda}    &     79.9      &      84.8      &  82.8   &  66.5   &    78.5       &5.9     \\
LMISA-3D~\cite{jafari2022lmisa_uda}    &84.5 & 82.8 & 88.6 & 70.1 &81.5 &2.3   \\
AllSpark~\cite{wang2024allspark}  &{87.0} &88.5 &86.4 &  \color{blue}\textbf{88.7} & 87.6 &{2.0} \\
A\&D~\cite{wang2024towards}  &\color{blue}\textbf{93.2} & 89.5 &\textbf{91.7} & 86.2 &\color{blue}{90.1} &\color{blue}{1.7} \\
DDSP~\cite{zheng2024dual}  & \color{red}\textbf{93.3} &\color{blue}\textbf{90.9} & 90.0 & 81.9 & 89.0 & 2.6 \\

\rowcolor{red!10} \textbf{Ours} &87.1 &\color{red}\textbf{94.3} &\color{blue}{90.9} &\color{red}\textbf{96.2} &\color{red}\textbf{92.1}{$\color{right}\textbf{\scriptsize{$\uparrow$}2.0}$}  &\color{red}\textbf{1.2}{$\color{right}\textbf{\scriptsize{$\downarrow$}0.5}$}  \\
\bottomrule

\bottomrule
\rowcolor{cyan!20}  \multicolumn{7}{c}{CT to MR} \\ \toprule
Vnet (Fully) &   82.8  &  80.5   & 92.4        & 78.8   &  83.6  &2.9 \\ \midrule
PnP-AdaNet~\cite{dou2019pnp_uda} &  43.7     & 68.9    & 61.9    & 50.8  &63.9    & 8.9  \\
% AdaOutput~\cite{tsai2018adaouput_uda}    &     60.8      &      39.8      &  71.5   &  35.5   & 51.9       &5.7   \\
CycleGAN~\cite{zhu2017cyclegan_uda}    &     64.3      &      30.7      &  65.0   &  43.0   &  50.7       &6.6   \\
CyCADA~\cite{hoffman2018cycada_uda}    &     60.5      &      44.0      &  77.6   &  47.9   &   57.5       &7.9  \\
% SIFA~\cite{chen2020sifa_uda}    &     65.3      &      62.3      &  78.9   &  47.3   &    63.4       &5.7\\
% DSFN~\cite{zou2020dsfn_uda}    &     84.7      &      76.9      &  79.1   &  62.4   &    75.8       &- \\
DSAN~\cite{han2021dsan_uda}    &     {71.3} &66.2 &76.2 &52.1 &66.5 &5.4     \\
LMISA-3D~\cite{jafari2022lmisa_uda}    &60.7 &72.4 & 86.2 &64.1 &70.8 & 3.6   \\
AllSpark~\cite{wang2024allspark}  & 72.7 &73.7 &85.2 &63.8 & 73.9 & 4.2 \\
A\&D~\cite{wang2024towards}  &62.8 &\color{red}\textbf{87.4} &61.3 &74.1 & 71.4 &7.9 \\
DDSP~\cite{zheng2024dual}  & \color{blue}\textbf{75.8} &\color{blue}\textbf{82.5} & \color{blue}\textbf{90.6} & \color{blue}\textbf{78.7} &\color{blue}\textbf{81.9} & \color{red}\textbf{2.3} \\

%/mnt/lee/Med/SSL_UDA/logs/Exp_UDA_MMWHS_ct2mr/diffusion_cutmask1_mt_v3_rec_corr_v2uda_300_1.0_15_1.0_0.1_0.1_0.9
\rowcolor{red!10}\textbf{Ours} &\color{red}\textbf{80.5} & 79.2 &\color{red}\textbf{93.3} &\color{red}\textbf{80.6} &\color{red}\textbf{83.4}{$\color{right}\textbf{\scriptsize{$\uparrow$}1.5}$} &\color{blue}\textbf{3.1}{$\color{right}\textbf{\scriptsize{$\uparrow$}0.8}$} \\
\bottomrule
\end{tabular}}
% \vspace{-0.4cm}
\end{table}

\begin{table*}[!htbp]
\centering
\footnotesize
\caption{Results on 2\% and 5\% labeled data settings of M\&Ms dataset for \textbf{Semi-MDG} task. 
% \textbf{Bold} and \underline{underline} denote the best and the second-best results.
}
% \vspace{-2mm}
\label{tab:mm}
\setlength\tabcolsep{2pt}
\centering
\resizebox*{1.0\textwidth}{!}{
\begin{tabular}{c|cccc|c||cccc|c}
\toprule
\multirow{2}{*}{Method}         & \multicolumn{5}{c||}{2\% Labeled data}  & \multicolumn{5}{c}{5\% Labeled data}\\
& Domain A        & Domain B        & Domain C & Domain D & \textbf{Average} & Domain A        & Domain B        & Domain C & Domain D & \textbf{Average} \\ \midrule
nnUNet~\cite{isensee2021nnunet} &   52.87  &  64.63   &  72.97   &    73.27  &  65.94  &   65.30  &  79.73   & 78.06        & 81.25   &  76.09     \\
SDNet+Aug~\cite{liu2021SDNet}     &54.48     &      67.81      & 76.46   &  74.35   &    68.28  &71.21    &77.31  &81.40     &79.95     &77.47                \\
LDDG~\cite{li2020lddg}    &     59.47      &      56.16      &  68.21   &  68.56   &    63.16    &     66.22      &      69.49      &  73.40   &  75.66   &    71.29        \\
SAML~\cite{liu2020saml}    &     56.31      &      56.32      &  75.70   &  69.94   &    64.57   &     67.11      &      76.35      &  77.43   &  78.64   &    74.88         \\
BCP~\cite{bai2023bidirectional}    &     71.57      &      76.20      &  76.87   &  77.94   &  75.65    &     73.66      &      79.04      &  77.01   &  78.49   & 77.05             \\
DGNet~\cite{liu2021dgnet}    &     66.01      &      72.72      &  77.54   &  75.14   &    72.85   &     72.40      &      80.30      &  82.51   &  83.77   &    79.75          \\
vMFNet~\cite{liu2022vmfnet}    &     73.13      &      77.01      &  81.57   &  82.02   &    78.43  &     77.06      &      82.29      &  84.01   &  85.13   &    82.12           \\
% AllSpark~\cite{wang2024allspark}    &83.77 &84.39 &85.29 &86.44 & 84.97 & 82.05 &84.39 &84.76 &84.27&86.36 
% \\
Meta~\cite{liu2021semi} & 66.01 &72.72 & 77.54 & 75.14 &72.85 &72.40 & 80.30 & 82.51 &83.77 & 79.75 \\
StyleMatch~\cite{zhou2023semi} & 74.51 &77.69 & 80.01 & 84.19 & 79.10 &81.21 &  82.04 &  83.65 &83.77 & 82.67 \\
EPL~\cite{yao2022enhancing}    & \color{blue} \textbf{82.35} &\color{blue}\textbf{82.84} &\color{red}\textbf{86.31} &\color{blue}\textbf{86.58} &\color{blue}\textbf{84.52} &\color{blue}\textbf{83.30} &85.04 &\color{red}\textbf{87.14} &\color{red}\textbf{87.38} &\color{blue}\textbf{85.72} \\
A\&D~\cite{wang2024towards}    & 79.62 & 82.26 & 80.03 & 83.31 & 81.31 & 81.71 &\color{blue}\textbf{85.44} & 82.18 & 83.90 & 83.31 
\\

\bottomrule
\rowcolor{red!10} \textbf{Ours} & \color{red}\textbf{83.09} & \color{red}\textbf{87.30} & \color{blue}\textbf{84.45} & \color{red}\textbf{87.02} & \color{red}\textbf{85.46}{$\color{right}\textbf{\scriptsize{$\uparrow$}0.94}$} & \color{red}\textbf{85.13} & \color{red}\textbf{87.87} & \color{blue}\textbf{84.90} & \color{blue}\textbf{86.98} & \color{red}\textbf{86.22}{$\color{right}\textbf{\scriptsize{$\uparrow$}0.5}$}  \\
\bottomrule

\end{tabular}}
\end{table*}

% \vspace{-4mm}
\subsection{Comparison with state-of-the-art methods}

\subsubsection{Comparison on SSMIS.} 

Tab.~\ref{tab:LA_4}, Tab.~\ref{tab:Sy_20}, Tab.~\ref{tab:Sy_10}, and Tab.~\ref{tab:amos} show quantitative comparison results for the LA, Synapse, and AMOS datasets. Our proposed DTLP-Net achieves notable enhancements in all evaluation metrics by a significant margin in different training scenarios, effectively leveraging the potential of unlabeled data. Specifically, on the LA dataset, our method achieves SOTA segmentation performance, surpassing the fully supervised method by 0.16\% (91.63\% vs. 91.47\%) and 0.96\%(92.43\% vs. 91.47\%)  in dice performance with only 5\% and 10\% labeled data, respectively. Our method achieves notable enhancements in Dice, Jaccard, 95HD, and ASD metrics, surpassing the second-best A\&D~\cite{wang2024towards}  performance by 1.7\%, 2.77\%, 0.85, and 0.45, respectively, for the 5\% setting. Similarly, with  10\% labeled data for training, we outperform the second-best approach GapMatch~\cite{huang2025gapmatch} by  1.43\% and 0.16 in Dice and ASD performance, respectively. These results are obtained without conducting any post-processing, ensuring fair comparisons with other methods.
 
Notably, on the Synapse dataset, our method achieves SOTA segmentation performance on most types of organs, which shows the promising ability to solve traditional SSMIS tasks, as shown in Tabs.~\ref{tab:Sy_20} and~\ref{tab:Sy_10} . Furthermore, by incorporating the class-imbalanced designs from~\cite{wang2024towards}, our method successfully segments the minor classes RAG and LAG, resulting in an improved overall Dice score of 66.01\% under the 20\% labeled setting. This achievement demonstrates a significant improvement (5.13\% in Dice) compared to previous methods.

We further evaluate our method on the AMOS dataset, and the experimental results are shown in Tab.~\ref{tab:amos}. Similarly, our method also achieves SOTA performance on the AMOS dataset. Specifically, as shown in Tab.~\ref{tab:amos}, on the  AMOS dataset with 2\% labeled data, our method outperforms the second-best method, \textit{i.e.}, AllSpark~\cite{wang2024allspark} with 8.79\% in Dice and 4.07 in ASD.

\begin{figure}[!htbp]
	\begin{center}
		\centering
		\includegraphics[width=.5\textwidth]{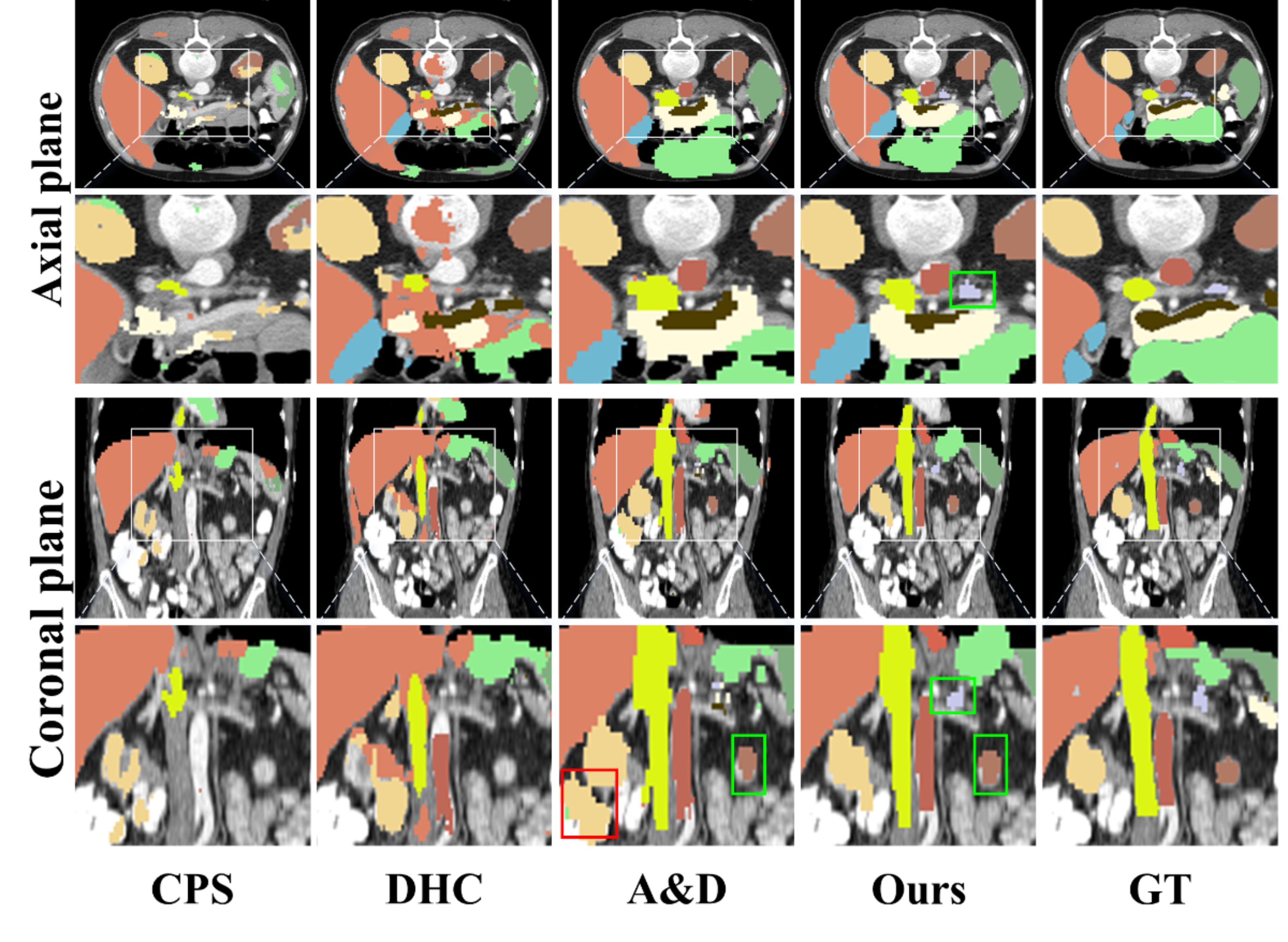}
		\centering
        % \vspace{-5mm}
		\caption{Visualization results with 20\% labeled on the Synapse dataset.}
        % \vspace{-5mm}
		\label{fig:synapse}
	\end{center}
\end{figure}

In addition, we also present qualitative results in Figs. \ref{fig:la} and ~\ref{fig:synapse}, demonstrating that our method delivers more accurate and smooth segmentation predictions compared to other methods.

\subsubsection{Comparison on UMDA.} 
Table \ref{tab:UMDA} presents the outcomes of our method for the UMDA task on the MMWHS dataset. In the Dice metric on MR to CT task, our methods outperform the fully-supervised method by 1.2\% (92.1\% vs. 90.9\%). Moreover, from CT to MR task, our methods are merely 0.2\% inferior to the fully-supervised method, showing its extraordinary performance. Compared to  UMDA methods that adopt image-level (CycleGAN~\cite{zhu2017cyclegan_uda}), feature-level (DSAN~\cite{han2021dsan_uda}) or both (CyCADA~\cite{hoffman2018cycada_uda},  DDSP~\cite{zheng2024dual} ) alignments to mitigate domain shifts, our method achieves comparable performance, surpassing the second-best DDSP~\cite{zheng2024dual}  performance by 3.1\% (92.1\% vs. 89.0\%) and 1.5\% (83.4\% vs. 81.9\%) in Dice on two  UMDA tasks. Furthermore, when compared with the generic semi-supervised segmentation method, \textit{e.g.}, A\&D~\cite{wang2024towards}, our method outperforms by 12.0\% (83.4\% vs. 71.4\%) in Dice on the CT to MR task.

\subsubsection{Comparison on Semi-MDA.} 

Tab.~\ref{tab:mm}  shows the comparison results of Semi-MDG methods on M\&Ms dataset. Compared to EPL~\cite{yao2022enhancing}, which adopts  Fourier transformation~\cite{yao2022enhancing} to deal with the domain shift, our methods achieve an improvement of 0.94\% (85.46\% vs. 84.52\%) and 0.5\%  (86.22\% vs. 85.72) with 2\% and 5\% labeled data, respectively. Notably, when compared with other SOTA pure semi-supervised segmentation methods, \textit{e.g.}, BCP~\cite{bai2023bidirectional}, our method shows solid performance gains of 9.81\% (85.46\% vs. 75.65\%) and  9.17\%(86.22\% vs. 77.05\%), respectively.  Furthermore,  we surpass the generic semi-supervised segmentation method  A\&D~\cite{wang2024towards}  by  4.15\% (85.46\% vs. 81.31\%) and  2.91\% (86.22\% vs. 83.31\%), respectively.  By leveraging the global - local data structure and devising an effective teaching strategy, our method generates more reliable pseudo - labels and mitigates the domain shift.

\begin{table}[!htbp]
\renewcommand\arraystretch{0.99}
\caption{Results of the ablation experiments carried out on the LA dataset, employing 5\% labeled and 10\% data.}
% \vspace{-2mm}
\resizebox{1\linewidth}{!}{
\begin{tabular}{ccccc|cccc}
\hline
$\mathcal{L}_{mix}$ & $\mathcal{L}_{mic}$ & $\mathcal{L}_{kd}$ & $\mathcal{L}_{rec}$ & $\mathcal{L}_{corr}$ & Dice
$\uparrow$ & Jaccard $\uparrow$ & 95HD $\downarrow$ & ASD $\downarrow$ \\ \hline
$\checkmark$ &  &    &   &  & 90.35 & 82.47 & 4.90 & 1.75   \\
$\checkmark$ &  $\checkmark$  &   &    &  & 90.94 & 83.43 & 4.42 & 1.60   \\
$\checkmark$ & $\checkmark$   &  $\checkmark$  &   &   & 90.73 & 83.09 & 4.76  & 1.60   \\
$\checkmark$ &  $\checkmark$ & $\checkmark$   & $\checkmark$    & & 91.16 & 83.80 & 4.46 & 1.59 \\
% $\checkmark$ & $\checkmark$   & $\checkmark$   &  $\checkmark$  & $\checkmark$   & \textbf{91.85} & \textbf{84.96}& \textbf{4.14} & \textbf{1.44}   \\ \hline
$\checkmark$ & $\checkmark$   & $\checkmark$   &  $\checkmark$  & $\checkmark$   & \textbf{91.63} & \textbf{84.59}& \textbf{4.40} & \textbf{1.41}   \\ \hline
$\checkmark$ &  &    &   &  & 91.61 & 84.58 & 4.86 & 1.54   \\
$\checkmark$ &  $\checkmark$  &   &    &  &  92.27 & 85.69 & 4.20 & 1.27  \\
$\checkmark$ & $\checkmark$   &  $\checkmark$  &   &   & 92.34 &85.50 & 4.13 & 1.30 \\
$\checkmark$ &  $\checkmark$ & $\checkmark$   & $\checkmark$    & & 92.24 & 85.63 & 4.32 & 1.26  \\
% $\checkmark$ & $\checkmark$   & $\checkmark$   &  $\checkmark$  & $\checkmark$   & \textbf{92.28} & \textbf{85.70}& \textbf{4.12} & \textbf{1.31}   \\ \hline
$\checkmark$ & $\checkmark$   & $\checkmark$   &  $\checkmark$  & $\checkmark$   & \textbf{92.43} & \textbf{85.96}& \textbf{3.94} & \textbf{1.30}   \\ \hline
\end{tabular}}
\label{tab:ablation_la}
% \vspace{-2mm}
\end{table}

\begin{table}[!htbp]
\renewcommand\arraystretch{0.99}
\caption{Results for ablation experiments which are conducted on the Synapse dataset using 20 \% labeled data.}
% \vspace{-2mm}
\resizebox{1\linewidth}{!}{
\begin{tabular}{ccccc|cc}
\hline
$\mathcal{L}_{mix}$ & $\mathcal{L}_{mic}$ & $\mathcal{L}_{kd}$ & $\mathcal{L}_{rec}$ & $\mathcal{L}_{corr}$ & Dice
$\uparrow$ & ASD $\downarrow$ \\ \hline
% $\checkmark$ &  &    &   &  & 71.52 & 5.41   \\
% $\checkmark$ &  $\checkmark$  &   &    &  & 81.28        & 2.67   \\
% $\checkmark$ & $\checkmark$   &  $\checkmark$  &   &   & 82.99  & 1.35   \\
% $\checkmark$ &  $\checkmark$ & $\checkmark$   & $\checkmark$    & & 83.40 & 1.91  \\
% $\checkmark$ & $\checkmark$   & $\checkmark$   &  $\checkmark$  & $\checkmark$   & \textbf{91.85} & \textbf{1.44}   \\ \hline
$\checkmark$ &  &    &   &  & 64.45$\pm$0.78 & 1.34$\pm$0.07  \\
$\checkmark$ &  $\checkmark$  &   &    &  & 64.58$\pm$1.17  & 1.40$\pm$0.07   \\
$\checkmark$ & $\checkmark$   &  $\checkmark$  &   &   & 65.45$\pm$0.12 & 1.29$\pm$0.10    \\
$\checkmark$ &  $\checkmark$ & $\checkmark$   & $\checkmark$    & &  65.27$\pm$0.76 & 1.28$\pm$0.09  \\
$\checkmark$ & $\checkmark$   & $\checkmark$   &  $\checkmark$  & $\checkmark$   & \textbf{66.01$\pm$0.43} & \textbf{1.69$\pm$0.79}   \\ \hline
\end{tabular}}
\label{tab:ablation_sy}
% \vspace{-2mm}
\end{table}

\subsection{Ablation Study}

\subsubsection{Effects of different components}

To validate the effectiveness of each component of our method in Eq.~\ref{eq:total},  i.e., global-local consistency loss $\mathcal{L}_{mix}$ and $\mathcal{L}_{mic}$, knowledge distillation loss $\mathcal{L}_{kd}$,  masked image reconstruction $\mathcal{L}_{rec}$,  and voxel-level label propagation loss $\mathcal{L}_{corr}^u$, we conduct ablation studies on the  LA dataset across two distinct semi-supervised configurations and Synapse dataset with 20\% labeled data for training. The outcomes of these ablation experiments are meticulously documented in Table \ref{tab:ablation_la} and \ref{tab:ablation_sy}, respectively.

\paragraph{ \textbf{Effectiveness of  Global-Local Consistency Learning (GLCL)}} To fully explore the data structure, except for cross-set CutMix Strategy that engenders novel training samples that fill the void between input samples to regularize the global distributional smoothness~\cite{li2024diversity}, we further adopt masking image modeling to learn the local semantics of the data.
As illustrated in Tab.~\ref{tab:ablation_la} and \ref{tab:ablation_sy}, it is apparent that GLCL  results in improved model performance.  For instance, from  Table \ref{tab:ablation_la}, by further adopting masking image modeling,  the Dice performance improves by 0.59\% (90.94 vs. 90.35\%) and 0.66\% (92.27\% vs. 91.61\%) with 5\%  and 10\% labeled data for training, respectively. Both cross-set CutMix and masking image modeling can reduce the domain shifts and improve the data-level diversity, thus improving the model generalization ability. Similar conclusions can be drawn from Table \ref{tab:ablation_sy}. However, compared with the LA dataset, the Synapse dataset challenges with class imbalance  and contains many small foregrounds, random masking may mask out the foregrounds, leading to suboptimal performance.

\paragraph{\textbf{Effectiveness of  knowledge distillation and masked reconstruction}} 

To further mitigate the potential noise in the hard pseudo-labels generated through Eq.~\ref{eq:pseudo_label}, we perform knowledge distillation from the Decoder branches $\mathcal{D}(x^l;\psi)$ and $\mathcal{D}(x^l,x^u;\xi)$ to the $\mathcal{D}(x^u;\theta)$ decoder, respectively. As can be clearly observed from Tables~\ref{tab:ablation_la} and \ref{tab:ablation_sy}, knowledge distillation is more effective on the Synapse dataset. However, it exhibits relatively minor improvements on the LA dataset. The underlying reason is that the Synapse dataset exhibits class imbalance and numerous small foregrounds. This results in the hard pseudo-labels prior being more prone to noise. However, distilling soft predictions from the other two branches may mitigate this issue somewhat. In addition,  the reconstruction of the features of the masked input volumes promotes contextual consistency, thereby enhancing the segmentation performance. As presented in Tables~\ref{tab:ablation_la}, through the additional implementation of masked reconstruction, the Dice performance increases from 90.73\% to 91.16\%.

\paragraph{\textbf{Effectiveness of voxel-level label propagation}} 
 To comprehensively explore the potential of unlabeled data, we investigate the label propagation strategy to capture the voxel-level pairwise similarities, thereby improving the segmentation performance.  As can be observed from Tables~\ref{tab:ablation_la} and \ref{tab:ablation_sy}, the Dice performance can be significantly enhanced.

\subsubsection{Effectiveness of Diverse teaching} 
To generate reliable pseudo labels and and realize diverse consistency learning, we employ the entropy-based teacher ensemble to acquire an ensembled prediction. As presented in Tab.~\ref{tab:ablation_tea}, when compared to using only one of the two teacher predictions, the ensembled diverse teaching can notably boost the model performance. Particularly on the Synapse dataset, this approach results in an approximately 2\% increase in the Dice score.

\begin{table}[!htbp]
\centering
\footnotesize
\renewcommand\arraystretch{0.99}
\caption{Effectiveness of Diverse teaching on the LA dataset with 5\% labeled data and Synapse dataset using 20 \% labeled data.}
\vspace{-2mm}
\resizebox{0.8\linewidth}{!}{
\begin{tabular}{cc|cc|cc}
\toprule
\multirow{2}{*} {$\mathcal{T}_{1}$} & \multirow{2}{*} {$\mathcal{T}_{2}$ } & Dice % \multirow{2}{*}{$\beta$} & \multicolumn{2}{c|}{Scans used} 
$\uparrow$ & ASD $\downarrow$ & Dice
$\uparrow$ & ASD $\downarrow$  \\ % \hline
\cline{3-6}
 & & \multicolumn{2}{c|} {LA (5\%)} & \multicolumn{2}{c}{Synapse (20\%) } \\ \hline
$\checkmark$ &  & 90.80 & 1.60  & 64.39$\pm$0.22 & 1.82$\pm$0.81 \\
 &  $\checkmark$  & 90.61  & 1.58   & 64.22$\pm$0.33 & 1.28$\pm$0.05 \\
$\checkmark$ & $\checkmark$   & \color{red}\textbf{91.63} & \color{red}\textbf{1.41}   & \color{red}\textbf{66.01$\pm$0.43} & \color{red}\textbf{1.69$\pm$0.79} \\ 
\bottomrule
\end{tabular}
}
\label{tab:ablation_tea}
\vspace{-0.8em}
\end{table}

\begin{figure}[!htbp]
    \centering

    \subfigure[Sensitivity to $\alpha$]{
        \begin{minipage}[t]{0.45\linewidth}
            \centering
            % \label{fig:stability.sub.1}
   \includegraphics[width=4.2cm]{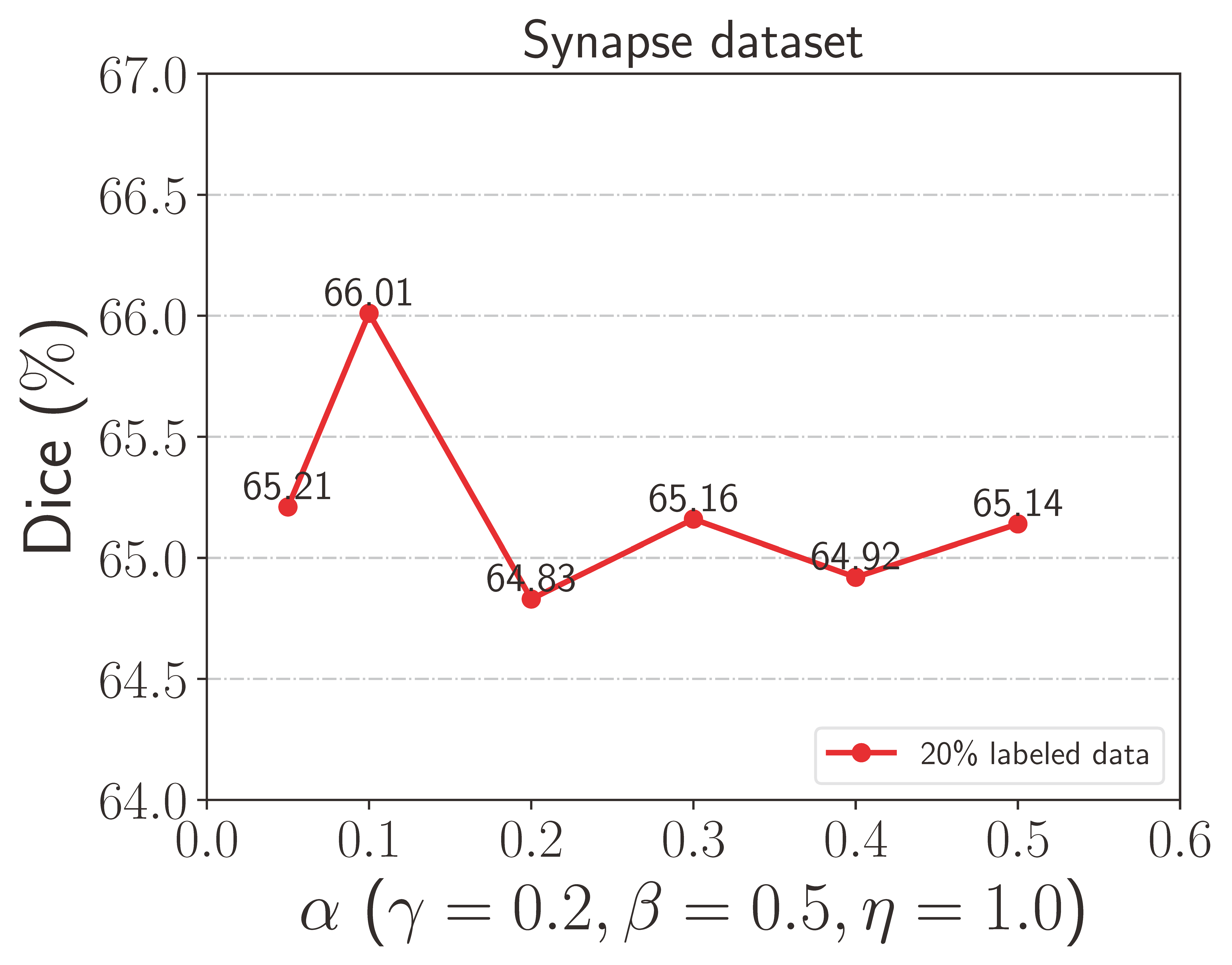}
        \end{minipage}
    }
    \subfigure[Sensitivity to $\beta$]{
        \begin{minipage}[t]{0.45\linewidth}
            \centering
            % \label{fig:stability.sub.2}
   \includegraphics[width=4.2cm]{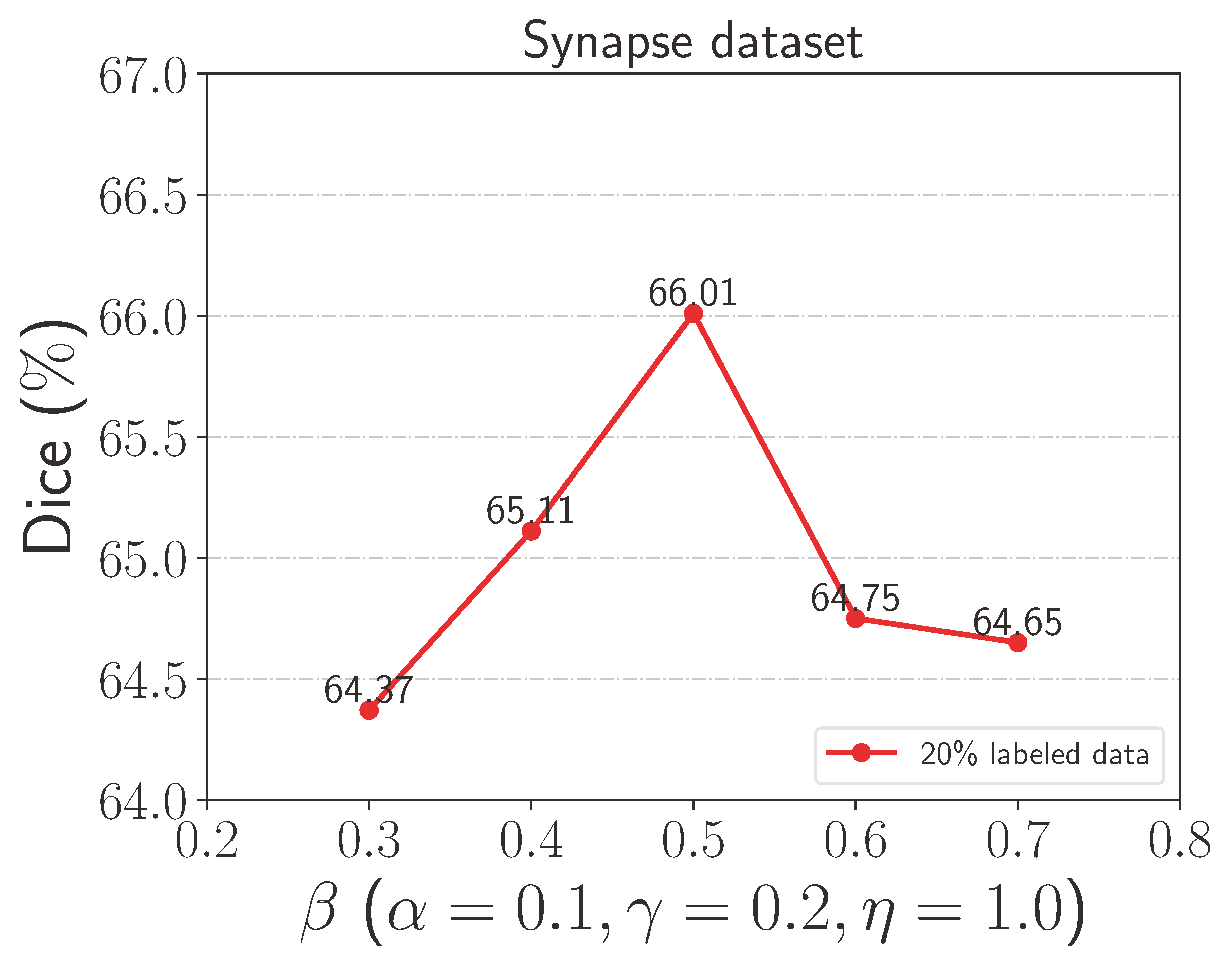}
        \end{minipage}
    }
    \subfigure[Sensitivity to $\gamma$]{
        \begin{minipage}[t]{0.45\linewidth}
            \centering
            % \label{fig:stability.sub.3}
   \includegraphics[width=4.2cm]{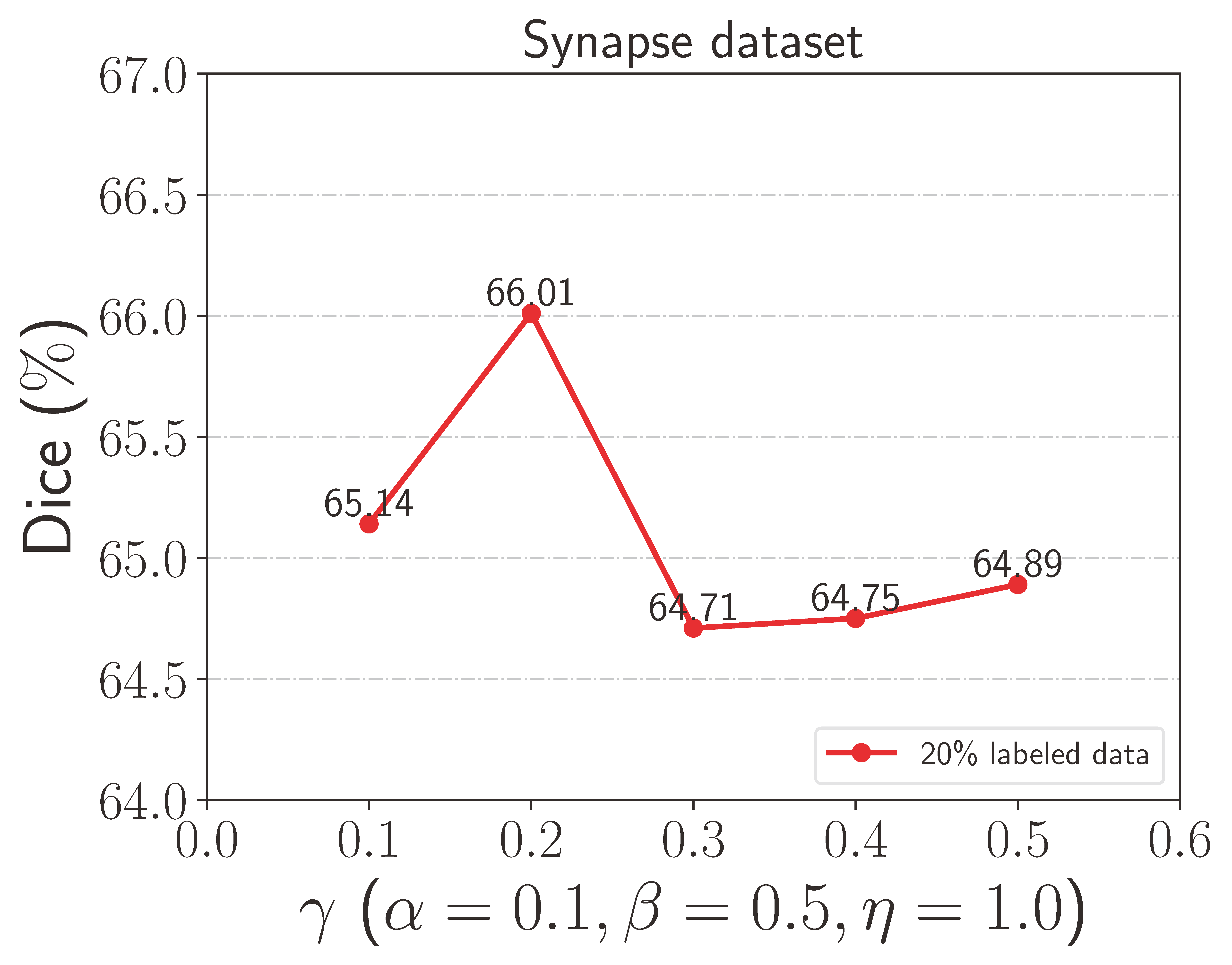}
        \end{minipage}
    }
    \subfigure[Sensitivity to $\eta$]{
        \begin{minipage}[t]{0.45\linewidth}
            \centering
            % \label{fig:stability.sub.4}
   \includegraphics[width=4.2cm]{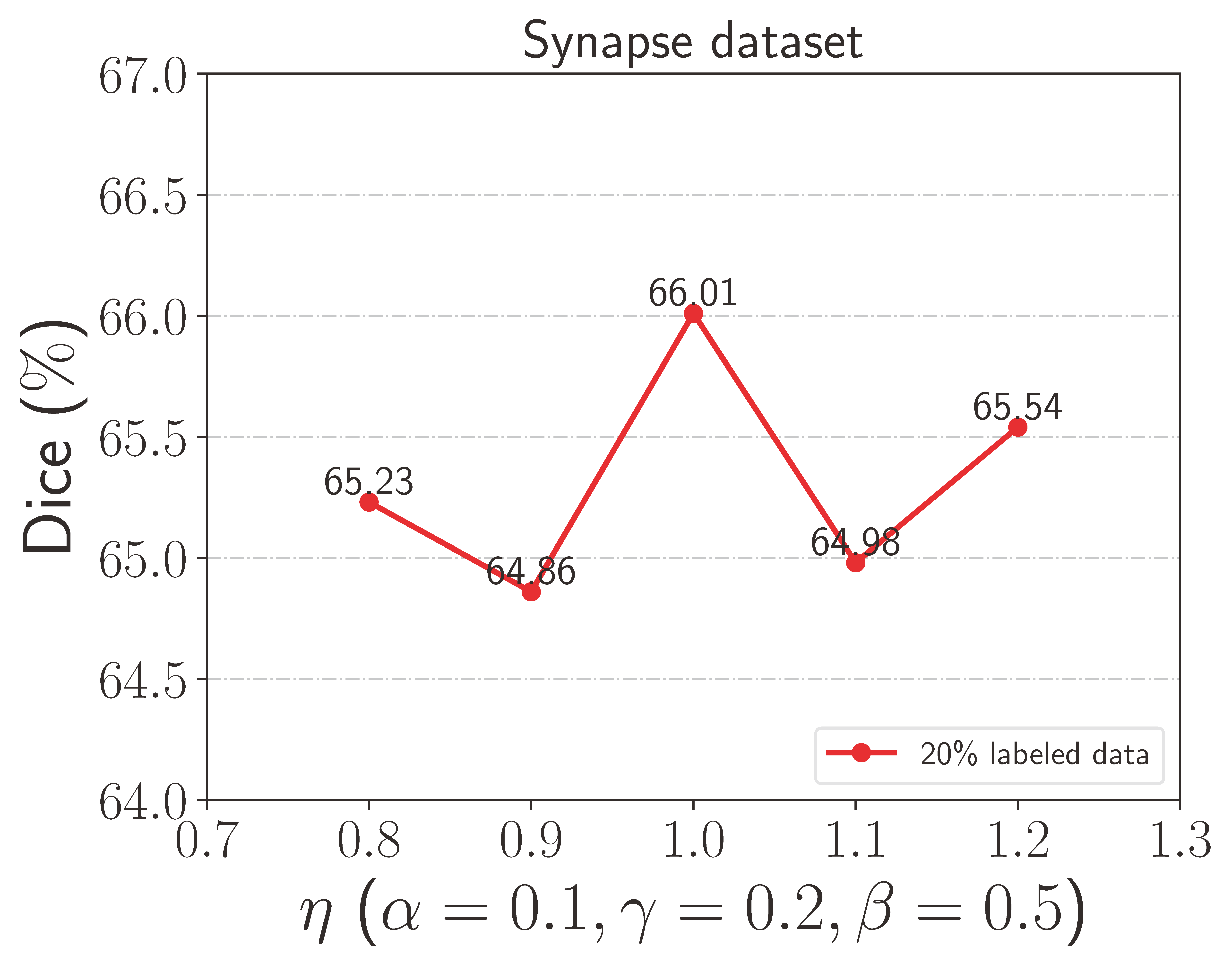}
        \end{minipage}
    }
    \caption{Parameters sensitivity analysis on Synapse data with 20\% labeled data for training.}
    \label{fig:sensitivity}
\end{figure}

% \vspace{-3mm}
\subsection{Further Analysis}

\subsubsection{Hyper-parameters analysis}
%% 参数敏感性分析
We further conduct ablation experiments on the Synapse dataset with 20\% labeled data to analyze the parameter sensitivity when training the network with various values  $\alpha$, $\beta$, $\gamma$, and $\eta$ in eq.~\ref{eq:total}. In general, as illustrated in Fig.~\ref{fig:sensitivity}, parameters that are either too high or too low may result in sub-optimal performance. Specifically, when utilizing masked images for consistency learning and reconstruction, $\alpha$ and $\gamma$ tend to assume relatively small values, where  $\alpha=0.1$ and $\gamma=0.2$ result in  superior performance. This is because the random masking strategy more readily masks small foregrounds, thereby misleading the model during training. Regarding the parameter $\gamma$, which governs the weight of knowledge distillation, a relatively small value of $\gamma$ causes the hard pseudo-labels to introduce noise. Conversely, a relatively high value of $\gamma$ results in noise being generated by the distilled soft predictions. Ultimately, the Dice score displays relatively slight fluctuations within a narrow range, suggesting that they have a relatively minor impact on the results.

\subsubsection{Ablation Study of mask ratio}

%% 参数敏感性分析

We further conduct ablation experiments on the Synapse dataset with 20\% labeled data to analyze the parameter sensitivity of mask ratio $r$ in eq.~\ref{eq:mask}. As shown in Tab.~\ref{tab:ablation_mask}, the value of mask ratio $r$ that is either too high or too low may result in sub-optimal Dice performance. A relatively high mask ratio has the potential to obscure the majority of the foreground organs, thereby misleading the model during the training process. On the other hand, a low value of the mask ratio $r$ gives rise to challenges in learning the domain-invariant features to reduce the distribution shifts and capturing the local context within the image.

\begin{table}[!htbp]
\footnotesize
\centering
\renewcommand\arraystretch{0.99}
\caption{Ablation Study of mask ratio on the Synapse dataset with 20\% labeled data.}
\vspace{-2mm}
\resizebox{0.7\linewidth}{!}{
\begin{tabular}{c|cc}
\toprule
mask ratio $r$  & Dice $\uparrow$ & ASD $\downarrow$ \\ \hline

0.4 & 65.41$\pm$0.35 & 1.34$\pm$0.08  \\
 0.5 &  \color{red}\textbf{66.01$\pm$0.43} & 1.69$\pm$0.79   \\
 0.6 & 65.23$\pm$0.34 & \color{red}\textbf{1.29$\pm$0.06}    \\
    % &  65.27$\pm$0.76 & 1.28$\pm$0.09  \\
  % & \textbf{66.01$\pm$0.43} & \textbf{1.69$\pm$0.79}   \\ 
\bottomrule
\end{tabular}}
\label{tab:ablation_mask}
\vspace{-0.8em}
\end{table}

% \vspace{-5mm}
\section{Conclusion}

This paper introduces a novel unified approach,  called Diverse Teaching and
Label Propagation Network (DTLP-Net), for addressing the annotation-efficient medical image segmentation tasks, including classical SSMIS, UMDA, and Semi-MDA. Our DTLP-Net incorporates diverse teaching strategies to generate reliable pseudo-labels for the student model. Subsequently, it realizes global-local consistency learning through inter-sample and intra-sample data augmentation, guided by these pseudo-labels. In addition, masked reconstruction on the feature level and knowledge distillation from the soft prediction is further utilized to alleviate the hard pseudo labels generated by the dual teachers. Ultimately, to fully exploit the structure of the data, a label propagation approach is put forward. This approach aims to learn pairwise similarities at the voxel-level for correlation consistency learning, thereby enhancing the model's generalization ability. Extensive experiments carried out on five benchmark datasets have verified the effectiveness of the proposed methodology, indicating the potential of our framework to tackle more challenging SSL scenarios.

\balance
\normalem
\bibliographystyle{IEEEtran}
\bibliography{arxiv}

\end{document}